%% file: iclr2027_conference.tex
\documentclass{article} 
\usepackage{iclr2027_conference,times}

\input{math_commands.tex}

\usepackage{hyperref}
\usepackage{url}
\usepackage{cleveref}
\usepackage{xspace}
\usepackage{todonotes}
\usepackage{booktabs}
\usepackage{enumitem}
\usepackage{amsthm,amsmath,amssymb}
\usepackage{tabularx}
\usepackage{booktabs}
\usepackage{float}
\usepackage{capt-of}

\newif\ifcomment
\commentfalse

\input{macros}

\ifcomment\newcommand{\riccardo}[1]{\textcolor{purple}{RA: #1}}
\else\newcommand{\riccardo}[1]{}\fi
\ifcomment\newcommand{\EK}[1]{\textbf{\textcolor{olive}{EK: #1}}}
\else\newcommand{\EK}[1]{}\fi
\ifcomment\newcommand{\AD}[1]{\textbf{\textcolor{violet}{AD: #1}}}
\else\newcommand{\AD}[1]{}\fi
\ifcomment\newcommand{\CG}[1]{\textbf{\textcolor{blue}{CG: #1}}}
\else\newcommand{\CG}[1]{}\fi
\ifcomment\newcommand{\MR}[1]{\textbf{\textcolor{red}{MR: #1}}}
\else\newcommand{\MR}[1]{}\fi
\ifcomment
\else\fi

\title{Neurosymbolic Learning of Interpretable World Models}

\author{Riccardo Andreoni \\
Fondazione Bruno Kessler\\
Free University of Bozen-Bolzano\\
\texttt{randreoni@fbk.eu} \\
\And
Massimiliano Ronzani \\
Fondazione Bruno Kessler \\
\texttt{mronzani@fbk.eu} \\
\AND
Chiara Ghidini \quad Alessandro Daniele\footnotemark[1] \\
Free University of Bozen-Bolzano \\
\texttt{chiara.ghidini@unibz.it} \\
\texttt{alessandro.daniele@unibz.it} \\
\And
\hspace*{10.6em}Emile van Krieken\footnotemark[1] \\
\hspace*{10.6em}Vrije Universiteit Amsterdam \\
\hspace*{10.6em}\texttt{e.van.krieken@vu.nl} \\
}

\iclrfinalcopy 
\begin{document}

\maketitle
\begingroup
\renewcommand{\thefootnote}{\fnsymbol{footnote}}
\footnotetext[1]{Equal supervision.}
\endgroup

\begin{abstract}
World models aim to learn compact representations of an environment from observations, enabling agents to predict future states and plan their behaviour. 
However, most existing approaches learn black-box representations, with limited ability to incorporate prior knowledge about the world. 
We introduce Concept-Grounded World Models (\methods), a neurosymbolic world model design that instead learns interpretable concept representations. 
We learn these concepts from observations using explicit and complete knowledge, expressed in temporal logic, that specifies the valid world states and the temporal relations governing their dynamics.
Furthermore, we directly use this knowledge by formulating planning as differentiable reasoning, optimising the agent’s actions by gradient ascent to satisfy a goal. 
We experimentally compare \method with a purely neural baseline and a neurosymbolic baseline on two game environments. 
We find that \methods ground concepts much more accurately, even over long horizons, resulting in significantly more robust planning both in and out of distribution.

\end{abstract}


\input{sections/introduction}
\begin{figure}
  \centering
  \includegraphics[trim={.5cm 1.2cm 0.5cm .0cm}, width=\linewidth]{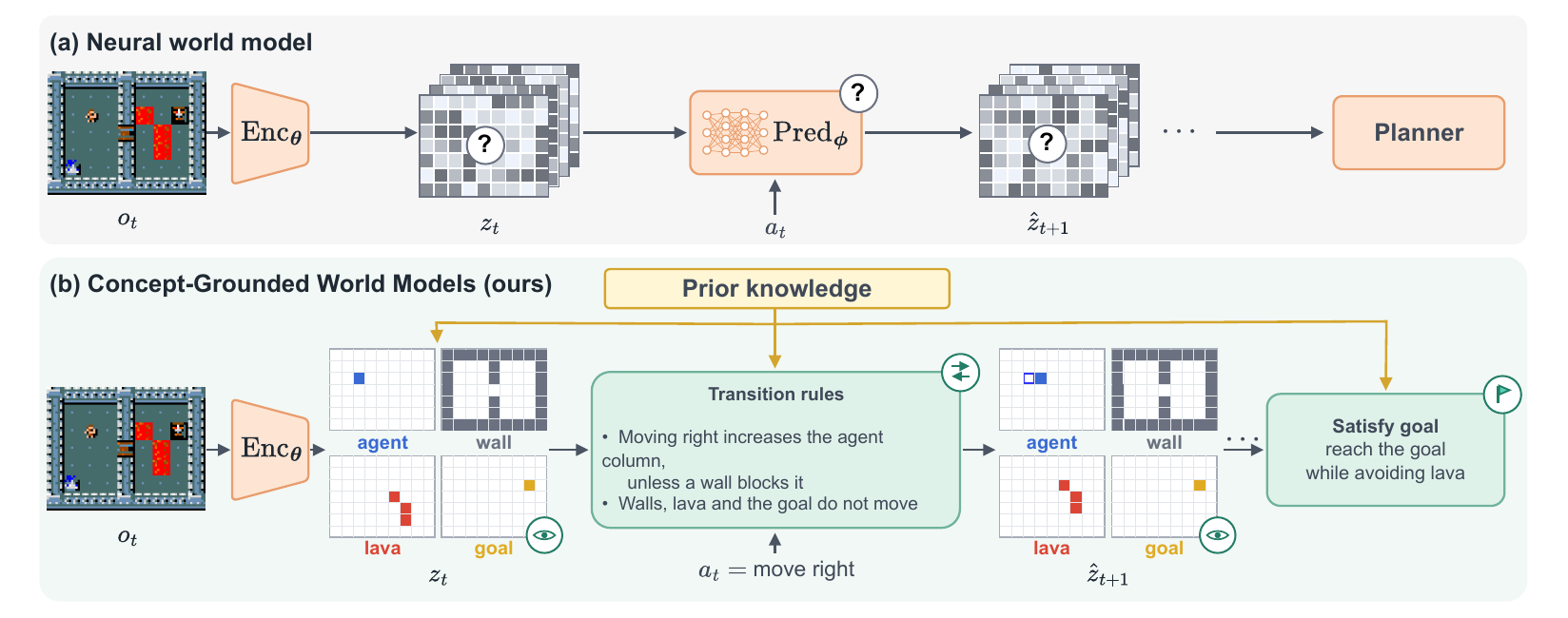}
  \caption{CGWMs are interpretable world models that ground concepts in images and use a symbolic transition system. This improves data efficiency, provides zero-shot transfer to new environments and allows for flexible and safe planning for new goals.
 }
  \label{fig:fig1}
\end{figure}
\input{sections/background}

\input{sections/method}
\input{sections/evaluation}
\input{sections/results}

\input{sections/related_works}
\input{sections/conclusions}
\section{AI Statement}
In this work, we used generative AI tools to implement part of the method, the code that generates the synthetic datasets, to aid writing part of the paper, to provide feedback on drafts, and to edit it for readability.
We have not used generative AI tools to develop theoretical models or conceptual frameworks, propose or refine hypotheses, design or provide feedback on research  methodology or experiments, clean and reformat dataset, support qualitative and thematic data analysis, interpret results, and the paper contains no mathematical claims or proofs.
The code, its outputs, and all the text drafted with AI assistance was reviewed and edited by the authors.
We take responsibility for the final content of this work, including text, claims or artifacts produced with the aid of generative AI.

\bibliography{iclr2027_conference}
\bibliographystyle{iclr2027_conference}

\appendix
\crefalias{section}{appendix}
\crefalias{subsection}{appendix}

\input{sections/appendix_ltlf}

\input{sections/appendix_envs}
\input{sections/appendix_results}

\input{sections/appendix_implementation}

\end{document}

%% file: math_commands.tex
\usepackage{amsmath,amsfonts,bm}

\def\eqref#1{equation~\ref{#1}}

\def\1{\bm{1}}

\DeclareMathAlphabet{\mathsfit}{\encodingdefault}{\sfdefault}{m}{sl}
\SetMathAlphabet{\mathsfit}{bold}{\encodingdefault}{\sfdefault}{bx}{n}



%% file: macros.tex
\newcommand{\method}{CGWM\xspace}
\newcommand{\methods}{CGWMs\xspace}

\newcommand{\DiffLTLf}{\ensuremath{\partial}LTL\ensuremath{_f}\xspace}
\newcommand{\nograd}[1]{\underline{#1}}
\newcommand{\Enc}{\mathrm{Enc}}
\newcommand{\Pred}{\mathrm{Pred}}

\newcommand{\LTL}{\textsc{ltl}\xspace}
\newcommand{\LTLF}{\textsc{ltl}\ensuremath{_f}\xspace}
\newcommand{\FLTLF}{\textsc{fltl}\ensuremath{_f}\xspace}

\newcommand{\tomorrow}{\mathsf{X}\xspace}
\newcommand{\weaknext}{\mathsf{X}_w\xspace}
\newcommand{\until}{\mathbin{\mathsf{U}}\xspace}
\newcommand{\always}{\mathsf{G}\xspace}
\newcommand{\eventually}{\mathsf{F}\xspace}
\newcommand{\release}{\mathbin{\mathsf{R}}\xspace}
\newcommand{\mrelease}{\mathbin{\mathsf{M}}\xspace}
\newcommand{\weakuntil}{\mathbin{\mathsf{W}}\xspace}
\newcommand{\limp}{\mathbin{\rightarrow}\xspace}
\newcommand{\U}{\until}
\newcommand{\X}{\tomorrow}
\newcommand{\Xw}{\weaknext}
\newcommand{\G}{\always}
\newcommand{\F}{\eventually}
\newcommand{\W}{\weakuntil}

\newcommand{\set}[1]{\{#1\}}
\newcommand{\prop}[1]{\mathtt{#1}}

\newcommand{\bo}{\mathbf{o}}
\newcommand{\vconcepts}{\mathbf{c}}
\newcommand{\concepttraj}{\mathbf{C}}
\newcommand{\concepts}{\mathcal{C}}
\newcommand{\concept}{c}
\newcommand{\cells}{\mathcal{G}}
\newcommand{\cell}{g}

\newcommand{\dottensor}{\mathbf{z}}
\newcommand{\dottraj}{\mathbf{Z}}
\newcommand{\btheta}{{\boldsymbol{\theta}}}

\newcommand{\p}{C\xspace}
\renewcommand{\P}{\concepts\xspace}

\newcommand{\timpl}{%
  \mathbin{
    \ooalign{
      \scalebox{0.78}{$\bigcirc$}\cr
      \hidewidth\raisebox{0.1ex}{$\mkern+.5mu\scriptscriptstyle\rightarrow$}\hidewidth\cr
    }
  }
}
\newcommand{\tand}{\otimes}
\newcommand{\tor}{\oplus}
\newcommand{\tnot}{\ominus}

\newcommand{\envfont}[1]{#1}
\newcommand{\IS}{\envfont{Ice Slider}\xspace}
\newcommand{\TR}{\envfont{Two Rooms}\xspace}
\newcommand{\TRe}{\envfont{Two Rooms (easy)}\xspace}
\newcommand{\TRm}{\envfont{Two Rooms (medium)}\xspace}
\newcommand{\TRh}{\envfont{Two Rooms (hard)}\xspace}
\newcommand{\TRld}{\envfont{Two Rooms (lava dense)}\xspace}
\newcommand{\TRpd}{\envfont{Two Rooms (periodic door)}\xspace}

%% file: sections/introduction.tex
\section{Introduction}
\label{sec:introduction}
One of the longstanding goals of artificial intelligence is to accurately model the world around us \citep{mccarthy1981some, DBLP:conf/icml/Sutton90}. 
This research has been reinvigorated since the deep learning revolution~\citep{DBLP:journals/corr/abs-1803-10122}, and especially in recent years, motivated by attempts to enable AI in the physical world, like robotics~\citep{DBLP:journals/csur/DingZSZZFYSLSXL26}. 
However, most of these approaches are purely neural, modelling transitions over opaque latent representations~\citep{DBLP:journals/corr/abs-2306-02572,DBLP:journals/corr/abs-2301-04104}.
Therefore, while world models are becoming more effective at predicting future states of the environment, they are difficult to inspect, adapt, and safely plan in.

We instead argue for a neurosymbolic world model design called Concept-Grounded World Models (\method) that uses symbolic transition systems acting on interpretable, abstract concepts~\citep{DBLP:journals/tmlr/KnabSBKSSSS26}. 
\method grounds these abstract representations in real-world data using an encoder neural network that bridges the continuous world with the symbolic concepts. 
After grounding, we can directly plan in the transition system with flexible goals that can include safety constraints. 
Furthermore, when the grounding of concepts from real-world data is robustly learned, \method can transfer zero-shot to new environments by merely adapting the transition system. 

Central to these benefits is learning a robust grounding of concepts. 
In this work, we study how learning from temporal knowledge and weak environment supervision may provide such groundings. 
For this, \method uses techniques from the field of neurosymbolic learning, where the goal is to ground concepts using a mix of weakly supervised data and background knowledge~\citep{DBLP:journals/corr/abs-2410-22077}. 
To model the kind of temporal knowledge required for world models, \method uses \FLTLF~\citep{DBLP:journals/corr/abs-2608-16443}, which relaxes \LTLF formulas with fuzzy semantics~\citep{DBLP:journals/ai/BadreddineGSS22,DBLP:journals/ai/KriekenAH22}. 
The fuzzy semantics ensure the differentiability of the \LTLF knowledge to enable direct integration with neural networks. 
To effectively minimise these losses, \method includes several training techniques to prevent poor local minima and ensure that concepts are properly grounded. 
Finally, in addition to training the encoder, we also apply \FLTLF to successfully plan in \methods. 


\textbf{Contributions}
%
(i) We introduce \method, a neurosymbolic world model design with a neural encoder that grounds a symbolic state of concepts, and a symbolic transition system encoded in \LTLF.
(ii) We develop algorithms for robust learning and planning in \methods using \FLTLF, a fuzzy relaxation of the \LTLF knowledge.
(iii) We experimentally compare \method with a purely neural baseline and a neurosymbolic baseline on two game environments\footnote{The code that reproduces all the experiments is available at \url{https://anonymous.4open.science/r/cgwm-726E}.}, and find that \method grounds concepts much more accurately, which results in improved robustness of planning both in and out of distribution, and that the choice of fuzzy semantics plays a crucial role in learning and planning.



%% file: sections/background.tex
\section{Background}
\label{sec:background}

\subsection{Linear Temporal Logic on Finite Traces}
\label{sec:background:LTLf}

The knowledge relevant to our setting is inherently temporal, we formalize it with temporal logics over traces.
In several application domains, such as planning, system dynamics are naturally captured by unbounded but \emph{finite} traces~\citep{DeDM14}.
This led to \emph{\LTL on finite traces} (\LTLF)~\citep{DBLP:conf/ijcai/GiacomoV13}. 
An \LTLF formula $\varphi$ over a set of $d$ concepts $\P$ extends propositional logic with temporal operators \emph{(strong) next} $\tomorrow$  and \emph{(strong) until} $\until$, and follows the syntax:
%
\begin{equation}
\varphi := \top \mid \bot \mid \p \mid \neg \varphi \mid \varphi \land \varphi \mid \varphi \lor \varphi \mid \varphi \to \varphi \mid \X \varphi \mid \varphi \U \varphi
\label{eq:syntax}
\end{equation}
where $\p\in \P$, the symbols $\top$ and $\bot$ are the \emph{true} and \emph{false} constants, respectively, and $\neg$, $\land$, $\lor$, $\to$ are Boolean connectives for
NOT, AND, OR, and implication.
The other temporal operators are defined in terms of these, e.g.,~$\eventually \varphi \equiv  \top \until \varphi$, and $\always \varphi \equiv \neg \eventually \neg \varphi$ for the \emph{eventually} and \emph{globally} operators.


The semantics of \LTLF is defined over sequences of time steps, each mapping \emph{every} concept to either $0$ (for \emph{false}) or $1$ (for \emph{true}). Formally, 
a \emph{trace} of length $n$ is a list of $d$-dimensional vectors $\concepttraj=\langle \vconcepts_0,\ldots,\vconcepts_{n-1}\rangle$ in which, for each time step $t = 0,\ldots,n-1$, the $t$-th vector $\vconcepts_t \in \{0, 1\}^{d}$ assigns a truth value to each concept $\p\in \P$. 

Intuitively, at time step $t$, $\tomorrow \varphi$ states that there is a next time step $t+1$ in which $\varphi$ holds; $\varphi_1 \until \varphi_2$ states that $\varphi_2$ holds in time step $t$ or a later time step $s>t$, and in all time steps in between, i.e., $t\leq t'<s$, $\varphi_1$ holds. 
$\eventually \varphi$ states that there is a future  step $t' \geq t$ in which $\varphi$ holds; and $\always \varphi$ states that $\varphi$ holds in all future steps $t' \geq t$. 
Appendix~\ref{appendix:ltlf} contains a complete formal introduction to \LTLF.

%

\subsection{Fuzzy Evaluation of Temporal Formulas}
\label{sec:background:fuzzy}
In this work, we employ \FLTLF to integrate temporal knowledge expressed in \LTLF into the learning process, and we evaluate formulas with \DiffLTLf, the neurosymbolic framework that implements this semantics~\citep{DBLP:journals/corr/abs-2608-16443}.
\FLTLF interprets \LTLF formulas with differentiable fuzzy logic operators to enable compatibility with gradient-based learning~\citep{DBLP:journals/ai/KriekenAH22}.
\FLTLF generalises the notion of trace defined above by relaxing the space of each truth vector $\vconcepts_t$ from Booleans $\{0,1\}$ to the interval $[0,1]$, which yields the fuzzy trace $\dottraj = \langle \dottensor_0, \dots, \dottensor_{n-1}\rangle$. 
Here, $z_{t, i}$ denotes the \emph{degree of truth} of a concept $\p_i \in \P$ at time step $t$.
%
We denote by $v(\varphi,\dottraj,t) \in [0,1]$ the degree of truth of an \FLTLF formula $\varphi$ over a fuzzy trace $\dottraj$ at time step $t \geq 0$.

Logical connectives $\lnot, \land, \lor, \to$ are evaluated by applying a fuzzy operator (namely, the negation $\tnot$, the t-norm $\tand$, the t-conorm $\tor$, and the implication $\timpl$, respectively) to the degrees of truth of their subformulas. 
For instance, $v(\varphi_1 \land \varphi_2,\dottraj,t) = v(\varphi_1,\dottraj,t) \tand v(\varphi_2,\dottraj,t)$.
Temporal operators like $\X$ and $\U$ instead relate the degree of truth at time step $t$ to those at subsequent time steps:
$v(\X \varphi,\dottraj,t) = v(\varphi,\dottraj,t{+}1)$ if $t < n-1$ and $0$ otherwise, and $v(\varphi_1 \U \varphi_2,\dottraj,t) = v(\varphi_2,\dottraj,t) \tor \bigl( v(\varphi_1,\dottraj,t) \tand v(\X(\varphi_1 \U \varphi_2),\dottraj,t) \bigr)$.


We write $v(\varphi,\dottraj) = v(\varphi,\dottraj,0)$ for the degree of truth of $\varphi$ in $\dottraj$. 
We employ two instances of the \FLTLF family: 
the G\"odel semantics, where $\alpha \tand \beta = \min(\alpha,\beta)$ and $\alpha \tor \beta = \max(\alpha,\beta)$, and the product semantics, where $\alpha \tand \beta = \alpha\beta$ and $\alpha \tor \beta = \alpha + \beta - \alpha\beta$.
Both use the standard negation $\tnot \alpha = 1-\alpha$ and interpret implication as the \emph{material implication} $\alpha \timpl \beta = (\tnot \alpha) \tor \beta$.


In this work, we primarily use the temporal operators \emph{Globally} ($\G$) and \emph{Eventually} ($\F$). 
To compute their degrees of truth, we use the following characterisation from~\cite{DBLP:journals/corr/abs-2608-16443}:
\begin{equation}
    v(\F \varphi,\dottraj) = \bigoplus_{t=0}^{n-1} v(\varphi,\dottraj,t),
    \qquad
    v(\G \varphi,\dottraj) = \bigotimes_{t=0}^{n-1} v(\varphi,\dottraj,t),
    \label{eq:F-G}
\end{equation}
where $\bigoplus$ and $\bigotimes$ denote the iterated application of the t-conorm $\tor$ and of the t-norm $\tand$, respectively.


%% file: sections/method.tex
\section{Concept-Grounded World Models}
\label{sec:method}

We introduce Concept-Grounded World Models (\methods), a type of neurosymbolic world model trained from \LTLF formulas (see \Cref{fig:fig1}). 
Existing neural world models use a purely neural latent state, and their features are not trained to have human-interpretable semantics~\citep{DBLP:journals/corr/abs-2306-02572,DBLP:journals/corr/abs-2301-04104}. 
Instead, our \methods represent latent states as truth values for each concept, enabling direct human interpretation.
We use prior knowledge about the environment, expressed in \LTLF over the same concepts, in three ways: to ground concepts in observations during training using only a small number of weak labels per trajectory; to replace the transition function of the world model; and to specify the goal at planning time.

\begin{figure}[t]
  \centering
  \includegraphics[trim={.0cm 0.5cm 2.5cm .0cm},  width=0.9\linewidth]{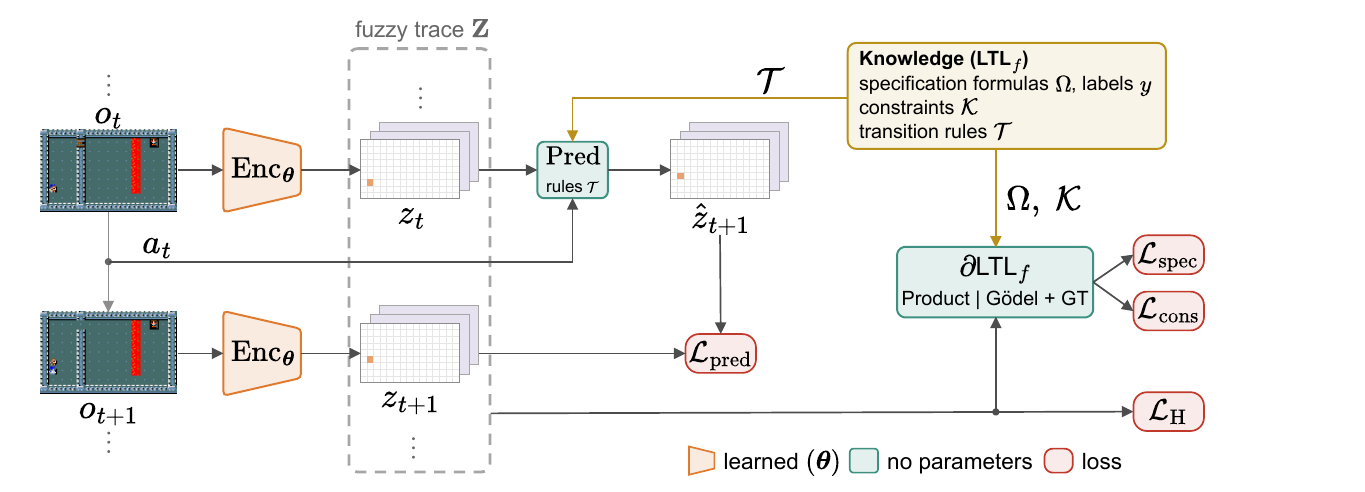}
  \caption{Overview of \method during learning.}
  \label{fig:method_learn}
\end{figure}

\subsection{Problem Setting}
\label{sec:method:setting}
We assume that, at each time step $t$, the environment provides an image $\bo_t$ and the agent executes an action $a_t\in \mathcal{A}$.
From this, we define a trajectory as a sequence $\tau = \langle \bo_0, a_0, \bo_1, \dots, a_{T-1}, \bo_T \rangle$ with $T\le T_{\max}$. 
In this paper, we assume that the environment is deterministic and fully observable, 
and that complete knowledge about the environment is available and exactly determines state transitions. 
We discuss extensions of this setting in \Cref{sec:conclusions}. 

A \method receives three types of rules about the environment as \LTLF formulas over concepts $\p$.
Constraints $\kappa \in \mathcal{K}$ represent rules that should hold on every trajectory, like $\always\, \neg(\prop{agent}_{1,2} \wedge \prop{wall}_{1,2})$.
Transition rules $\mathcal{T}$ describe transitions in concept space, defining the symbolic state at time step $t+1$ from the state and action at time step $t$, such as $\big(\prop{agent}_{1,2} \wedge \prop{right} \wedge \neg\, \prop{wall}_{1,3}\big) \rightarrow \tomorrow \prop{agent}_{1,3}$.
Finally,  we have a set $\Omega = \{\omega_1, \dots, \omega_M\}$ of specification formulas which, unlike constraints, hold only on some trajectories and not on others. 
One example is $\omega_m=\eventually\, \prop{at\_goal}$, which states that the agent is eventually in the same cell as the goal.
In particular, each training trajectory comes with a binary label $y_m$ per specification formula, equal to $1$ if $\concepttraj \models \omega_m$ and to $0$ otherwise.
These labels provide only weak supervision: we do not give labels for individual concepts $\vconcepts_t$.
Each formula is written for every cell, and \Cref{appendix:envs} lists the knowledge of each environment.


\subsection{Model}
\label{sec:method:model}
A \method consists of two components, an encoder and a predictor (\Cref{fig:method_learn}).
The encoder maps an image to a symbolic state, which assigns a degree of truth in $[0,1]$ to every concept $\p$.
The predictor computes the symbolic state at time step $t+1$ from the symbolic state and the action at time step $t$:
\begin{equation}
  \text{Encoder: } \dottensor_t = \Enc_\btheta(\bo_t) \in [0,1]^{d},
  \qquad
  \text{Predictor: } \hat{\dottensor}_{t+1} = \Pred(\dottensor_t, a_t).
  \label{eq:model}
\end{equation}
The encoder is a neural network with $d$ output channels, one per concept, followed by a sigmoid function, so that the entry $z_{t, i}$ is the degree of truth of concept $\p_i$ at time step $t$.
The predictor has no learnable parameters, and it applies the transition rules $\mathcal{T}$ to the current state and action, evaluated with \DiffLTLf (\Cref{sec:background:fuzzy}).
Since these rules are written over the same concepts $\P$, the predicted state is again an assignment of degrees of truth to those propositions.
Constraints and specification formulas are evaluated on a fuzzy trace $\dottraj$ in the same way.
We write $v(\omega, \dottraj)$ and $v(\kappa, \dottraj)$ for the degrees of truth of a specification formula $\omega$ and of a constraint $\kappa$ on a fuzzy trace $\dottraj$, which are both differentiable with respect to the encoder parameters $\btheta$.

\subsection{Learning}
\label{sec:method:learning}
Only the encoder parameters $\btheta$ are learned, with the loss
\begin{equation}
  \mathcal{L}(\btheta) = \mathcal{L}_{\text{spec}} + \mathcal{L}_{\text{cons}} + \mathcal{L}_{\text{pred}} + \lambda_{\text{H}} \, \mathcal{L}_{\text{H}}.
  \label{eq:loss}
\end{equation}

The \textbf{specification loss} $\mathcal{L}_{\text{spec}}$ is the binary cross-entropy between $v(\omega_m,\dottraj)$ and the label $y_m$, averaged over formulas and trajectories.
It is the only term that uses supervision, and it constrains the trace as a whole rather than any single proposition.
The \textbf{constraint loss} $\mathcal{L}_{\text{cons}}$ is the average of $-\log v(\kappa,\dottraj)$ over the constraints $\kappa \in \mathcal{K}$, and is minimised when every constraint is fully satisfied.
The \textbf{prediction loss} $\mathcal{L}_{\text{pred}}$ is the binary cross-entropy between the predicted next state $\hat{\dottensor}_{t+1}=\Pred(\Enc_\btheta(\bo_t), a_t)$ and the state $\dottensor_{t+1}=\Enc_\btheta(\bo_{t+1})$ encoded from the next observation.
Although the \method predictor does not have learnable parameters, it is differentiable, so gradients still reach the encoder parameters $\btheta$.
The \textbf{entropy term} $\mathcal{L}_{\text{H}}$ is the negated binary entropy of $\dottraj$, averaged over concepts and time steps.
Its weight $\lambda_{\text{H}}$ decreases linearly to zero during the first half of training.
Minimising $\mathcal{L}_{\text{H}}$ prevents concept truth values from prematurely saturating to $0$ or $1$ at the beginning of training, which we found helps to avoid poor local minima.

Many constraints $\kappa\in\mathcal{K}$ are implications of the form $\always\,(\alpha \rightarrow \beta)$. 
In an environment with an agent and walls, an example constraint 
is $\always\,\big(\prop{agent}_{1,2} \wedge \prop{right} \wedge \tomorrow\, \prop{agent}_{1,2}\big) \rightarrow \prop{wall}_{1,3}$: if the agent moves right and remains in the same cell, then the cell to its right is a wall. 
Here, the antecedent is about $\prop{agent}$ and is used as evidence, and the consequent is about $\prop{wall}$, the concept that receives the evidence.
A fuzzy implication can become true either by raising the degree of truth of the consequent $\beta$ (\emph{modus ponens}), or by lowering that of the antecedent $\alpha$ (\emph{modus tollens})~\citep{DBLP:journals/ai/KriekenAH22}.
By default, minimising $\mathcal{L}_{\text{cons}}$ enables both directions.
However, the modus tollens direction changes the concepts that provide the evidence, and it admits local minima in which the antecedent is false everywhere, satisfying the constraint trivially. 
In our example, the encoder can set $\prop{agent}$ to false in every cell. 
We therefore prevent gradients from flowing into the antecedent, which we denote by $\nograd{\alpha}$, and write the constraint as $\always\,(\nograd{\alpha} \rightarrow \beta)$.

\subsection{Planning}
\label{sec:method:planning}
\begin{figure}[t]
  \centering
  \includegraphics[trim={.0cm 0.5cm 2.5cm .0cm}, width=0.95\linewidth]{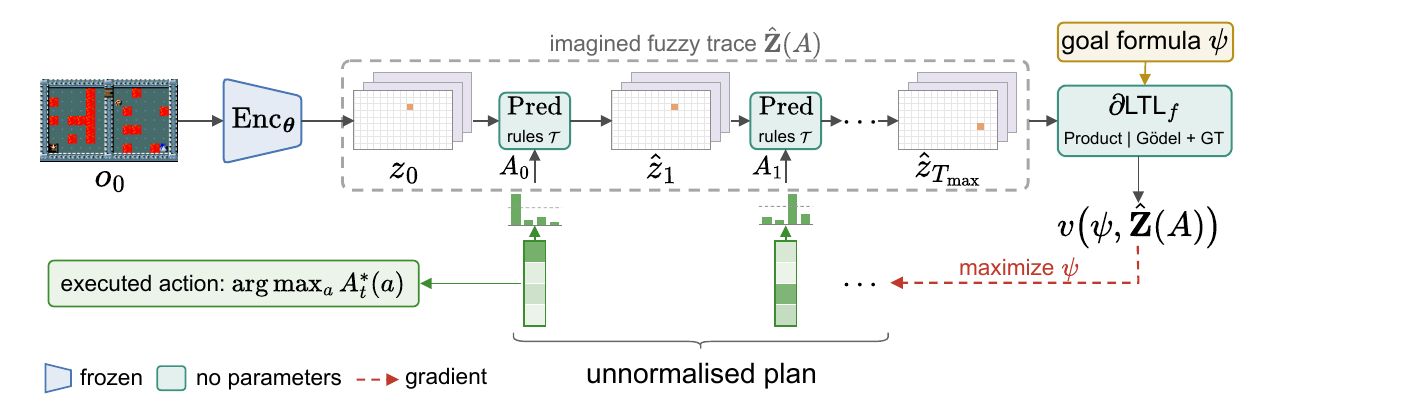}
  \caption{Overview of \method during planning.}
  \label{fig:method_plan}
\end{figure}

\EK{we should make clearer why we don't use classical planning and are using gradient ascent. I think Ale mentioned compatibility with hybrid states. Right now it's only mentioned in the conclusion}

After the world model has been trained, it can be leveraged to plan towards a goal, which can be expressed as an \LTLF specification $\psi$. 
Rather than relying on classical planning, we formulate planning as a differentiable optimisation problem to enable future hybrid world models with both symbolic and continuous variables (\Cref{sec:conclusions}).
When planning with \method, we freeze the encoder parameters and apply the encoder to the first image of the episode, $\hat{\dottensor}_0 = \Enc_\btheta(\bo_0)$ (\Cref{fig:method_plan}). 
Afterwards, no other observations are received and planning happens only on the imagined symbolic states.
A plan is parameterised by a matrix $U \in \mathbb{R}^{T_{\max} \times |\mathcal{A}|}$, whose rows $\mathbf{u}_t$ are transformed into action truth degrees $A_t \in [0,1]^{|\mathcal A|}$ according to the fuzzy semantics used for planning.
Starting from $\hat{\dottensor}_0 = \Enc_\btheta(\bo_0)$, the predictor uses $A_t$ in place of the discrete action $a_t$ and produces the imagined fuzzy trace $\hat{\dottraj}(A) = \langle \hat{\dottensor}_0, \dots, \hat{\dottensor}_{T_{\max}} \rangle$:
\begin{equation}
  \hat{\dottensor}_{t+1} = \Pred(\hat{\dottensor}_t, A_t), \qquad t = 0, \dots, T_{\max} - 1.
  \label{eq:rollout}
\end{equation}
We then plan by searching for matrices $A$ that maximise the degree of truth of the goal formula:
\begin{equation}
  A^\ast = \arg\max_{A} \; v\big(\psi, \hat{\dottraj}(A)\big).
  \label{eq:planning}
\end{equation}
Since the encoder is frozen and the predictor has no parameters, gradients only propagate to the matrix $U$.
We optimise \Cref{eq:planning} by gradient ascent, running several independent initialisations of $U$ in parallel.
We execute the best imagined plan by taking, at each time step $t$, the action $a_t$ with the highest degree of truth $\arg\max_{a_t} A_{t, a_t}$.

The encoder is used only to recognise the concepts in $\concepts$ in the first image, while transition rules $\mathcal{T}$ and goal formula $\psi$ specify how the world evolves and what the task is.
As a consequence, the same trained \method world model can plan in environments with different layouts or grid sizes, with different dynamics by replacing $\mathcal{T}$, and for different tasks by replacing $\psi$, as long as the set of concepts $\concepts$ is unchanged (\Cref{sec:evaluation}).

\subsection{Fuzzy Semantics for Learning and Planning}
\label{sec:method:semantics}
Which fuzzy semantics of \FLTLF (see \Cref{sec:background:fuzzy}) best suit learning and planning in \method? 
Product semantics smoothly propagates gradients through multiple operands and is generally well suited for learning, while the sparse gradients of Gödel semantics are often seen as inefficient for learning~\citep{DBLP:journals/ai/KriekenAH22}.
However, recent work on the Gödel Trick, which adds noise to the Gödel semantics, found that it can outperform product semantics on combinatorial problems~\citep{DBLP:conf/kr/DanieleK26}.
This suggests that the Gödel Trick may be particularly well suited for planning, which requires searching over discrete actions.

We test this hypothesis by evaluating both semantics in both phases, computing product semantics in logarithmic space for numerical stability~\citep{DBLP:journals/corr/abs-2306-14546}. During learning, we use the two semantics directly, without adding noise.
During planning, we add noise to explore different plans.
For Gödel semantics, we use the Gödel Trick and follow the categorical-variable construction of~\citet{DBLP:conf/kr/DanieleK26}. 
For product semantics, we instead use Gumbel-Softmax~\citep{DBLP:conf/iclr/JangGP17} to obtain the action truth values $A_t$.
Noise is sampled at every optimisation step to enable escaping local minima.
We report variants without noise in \Cref{tab:planning_ablation}.

%% file: sections/evaluation.tex
\section{Experimental Setup}
\label{sec:evaluation}
We evaluate \method on two grid-world environments with different dynamics and compare it with LeWorldModel (LeWM) \citep{DBLP:journals/corr/abs-2603-19312}, a purely neural world model, and NeSy-MM \citep{DBLP:conf/aaai/SmetVRM25}, a neurosymbolic Markov model that we adapt to our setting.
The experiments address four research questions.
\begin{description}[nosep]
    \item[\textbf{RQ1}:] Does \method learn to recognise the concepts in $\concepts$ from images using only trajectory labels and knowledge, while requiring fewer training trajectories than the baselines?
    \item[\textbf{RQ2}:] Is the grounding preserved in the imagined trace as the rollout horizon grows?
    \item[\textbf{RQ3}:] Can \method plan safely and effectively both in the environments it has been trained on and in environments with different layouts or dynamics?
    \item[\textbf{RQ4}:] How does the choice of fuzzy semantics affect learning and planning?
\end{description}

Next, we describe the environments, baselines, and protocols (more settings in 
\Cref{appendix:envs,appendix:implementation}).


\begin{figure}[t]
  \centering
  \includegraphics[trim={.0cm 0.5cm .0cm .0cm}, width=0.95\linewidth]{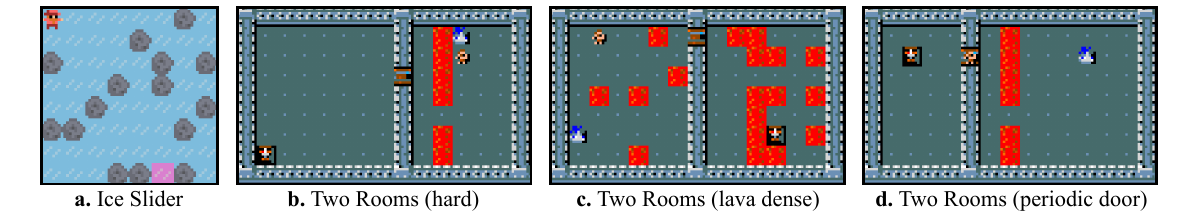}
  \caption{Initial observations of the environments: (a) \IS, (b) \TRh, (c) \TRld, and (d) \TRpd, where the door has closed on the agent.}
  \label{fig:envs}
\end{figure}

\paragraph{Environments.}
The first environment is \IS (\Cref{fig:envs}a)~\citep{DBLP:conf/nips/BagatellaORM21}, a puzzle in which the agent slides along the chosen direction until it hits a rock or the grid border, the episode ends when it stops on the goal.
Its concepts are $\concepts_{IS}=\{\prop{agent}, \prop{rock}, \prop{goal}\}$.
The second environment, \TR (\Cref{fig:envs}b), is built with MiniHack~\citep{DBLP:conf/nips/SamvelyanKKPJHP21} to reproduce a NetHack level~\citep{DBLP:conf/nips/KuttlerNMRSGR20}: a grid divided by a wall into two rooms, where the agent must reach the goal in the other room by first stepping on a button that opens the door, while avoiding lava, which ends the episode.
Its concepts are $\concepts_{TR}=\{\prop{agent},\prop{goal}, \prop{wall}, \prop{lava}, \prop{button},\prop{door}\}$, and it has richer dynamics than \IS.
We consider three variants of \TR of increasing complexity: 
\TRe has only the agent, the goal and the walls; \TRm adds the button and the closed door; and \TRh adds the lava.
The layouts are randomised in every episode for \IS (rocks and goal), \TRe (goal), \TRm (goal and button), and \TRh (inner wall, lava and all objects).
To test generalisation, we derive two environments from \TRh: \TRld (\Cref{fig:envs}c) keeps the dynamics but changes the layout, with three times as many lava cells, while \TRpd (\Cref{fig:envs}d) changes the dynamics: once the button is pressed, the door opens and closes at each time step, crushing any agent standing on the door cell as it closes.
\Cref{appendix:envs} describes all environments in detail, alongside used formulas.

\paragraph{Baselines.}
To evaluate the benefits of explicit symbolic states and of training on prior knowledge, we compare \method against LeWM, a JEPA world model that uses neural networks for both the encoder and the predictor~\citep{DBLP:journals/corr/abs-2603-19312}.
We give LeWM a latent vector for each cell of the grid, to match the spatial structure of \method.
We also give LeWM a reward at every frame (whereas \method only receives per-trajectory labels), for instance when the agent is at the goal or first presses the button, and a negative reward when it steps on lava.
LeWM learns to predict this reward from its latent state with a reward head. 
At planning time, LeWM optimises the actions by gradient descent on the predicted reward.
Unlike \method, which plans once from the first image, LeWM gets environment observations every five actions and re-plans.
Although these changes depart from the original LeWM, they all favour the baseline. 
We further motivate them in \Cref{appendix:implementation:baselines}. 


We also compare \method with NeSy-MM~\citep{DBLP:conf/aaai/SmetVRM25}, a neurosymbolic Markov model with a neural encoder and a predictor based on a symbolic probabilistic logic program, which makes it close to our assumptions.
However, \citet{DBLP:conf/aaai/SmetVRM25} either provide symbolic states, or only model the agent's position, with direct supervision on the agent's final position. 
We therefore give it the same concepts, knowledge, and trajectory labels as \method.
NeSy-MM does not come with a planning procedure, so we only compare on concept grounding.

\paragraph{Protocol and Metrics.}
We evaluate data efficiency by varying the number of training trajectories $N \in \{50, 200, 800, 3200\}$, using ten random seeds for each setting. 
We give LeWM approximately four times as many optimisation steps as \method to ensure convergence. 
All reported results are the mean and standard deviation over the ten seeds. 
We evaluate the learned world models along two dimensions. We measure the quality of the learned concept representations with the F1 score, and measure planning performance by the fraction of test instances successfully solved. 
Protocols specific to each evaluation are described alongside the corresponding results in \Cref{sec:results}.

%% file: sections/results.tex
\section{Results}
\label{sec:results}

\begin{figure}[H]
  \centering
  \includegraphics[trim={.0cm 0.5cm .0cm .5cm}, width=0.95\linewidth]
  {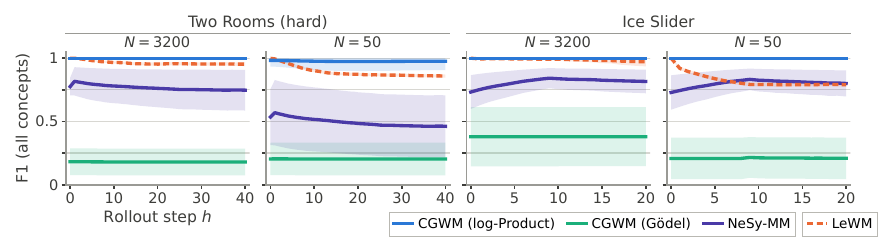}
  \caption{\textbf{\method with log-product, learns accurate grounding from weak supervision and knowledge.} F1 on the concepts along the imagined trace, for the largest and smallest training sets.
  }
  \label{fig:rq12}
\end{figure}

In this section, we answer the four research questions of \Cref{sec:evaluation}. 
The complete results are reported in \Cref{appendix:results}.
To answer RQ1 and RQ2, we measure concept recognition using the F1 score, both on states encoded from observations and along imagined rollouts of increasing horizon. Since LeWM has no explicit concepts, we train a linear probe on its latent states using the ground-truth concepts of the training trajectories. \Cref{fig:rq12} summarises the results.

\paragraph{RQ1: Grounding from observations.}
Trained with log-product semantics, 
\method learns an explicit symbolic state whose concepts are almost always grounded perfectly.
Its F1 remains perfect as the size of the training set decreases, with only slightly lower performance on the \TR environment with 50 trajectories (\Cref{fig:rq12}).
Instead, when trained with Gödel semantics, it does not ground the concepts at any training set size.
NeSy-MM, which receives the same knowledge and trajectory labels as \method, grounds only part of the concepts (more details in \Cref{appendix:results}), and its performance degrades with smaller training set sizes in the \TR environment. 
Additionally, \Cref{appendix:results:ablations} ablates the entropy loss term $\mathcal{L}_{\text{H}}$ and the blocking of modus tollens gradients (\Cref{sec:method:learning}). 
The ablations have poor grounding performance resulting from local minima. 
Finally, after training a linear probe from ground-truth concepts, LeWM recognises the concepts just as well as \method on single observations, but it differs when imagining the next steps.

\paragraph{RQ2: Long-horizon grounding.}
\method and NeSy-MM imagine next steps with the transition rules rather than with a learned predictor.
As a consequence, when the grounding is perfect, the predicted symbolic state should remain correct at every horizon, since the environments are deterministic. 
For \method, this is confirmed by the results on the longest rollouts (\Cref{fig:rq12}). 
NeSy-MM, instead, starts from an imperfect grounding and propagates initial errors to subsequent time steps. 
LeWM starts from a nearly perfect initial state, but its performance degrades with longer horizons. 
A simple explanation is that prediction errors accumulate by repeatedly applying the learned dynamics, whereas \method can apply the given transition rules exactly. 
This is supported by the results at smaller training set sizes, where LeWM’s performance degrades more rapidly with rollout steps, suggesting that the learned dynamics do not generalise accurately when trained on less data.


\EK{In the intro we promise 'safe planning', but here we only say we optimise towards going to the goal. It should be clear that we also evade the lava (it could be this is already included implicitly, but good to make explicit here)}

\paragraph{RQ3: Planning.} To answer RQ3, we evaluate \method and LeWM on the same 100 test instances per seed.
\method optimises the goal formula $\psi = \eventually\;\prop{at\_goal}$ on its imagined trace, whereas LeWM optimises its predicted reward.
Plans have a maximum horizon of $40$ actions in \TR and $20$ in \IS, and all test instances are solvable within these horizons.

In \TR, LeWM is trained separately on each of the three variants, while \method always uses the world model trained on \TRh.
We next evaluate generalisation without retraining on \TRld and \TRpd.
In the latter, \method only replaces the transition rules, while keeping its encoder unchanged.
Planning performance is measured as the fraction of test instances solved.
We note that if \method finds a plan, it is automatically safe, as the transition rules state that lava ends the episode, and so the agent fails to reach the goal. 
The results are summarised in \Cref{tab:planning}.
\begin{table}[t]
\caption{\textbf{\methods learn representations that allow for robust planning both in and out of distribution.} Percentage of planning instances solved in the environment, mean and standard deviation (in parentheses) over 10 seeds. For \method, the semantics used for planning is in parentheses.}
\centering
\footnotesize
\setlength{\tabcolsep}{3pt}
\label{tab:planning}
\begin{tabular}{lccccccc}
\toprule
 & \multicolumn{5}{c}{\TR} & \multicolumn{2}{c}{\IS} \\
\cmidrule(lr){2-6}\cmidrule(lr){7-8}
 & easy & medium & hard & lava dense & periodic door & $T{=}20$ & $T{=}T^*$ \\
\midrule
\method (G\"odel Trick) & {\bf 100.0\,{\scriptsize(0.0)}} & {\bf 100.0\,{\scriptsize(0.0)}} & {\bf 96.3\,{\scriptsize(0.8)}} & {\bf 88.4\,{\scriptsize(2.8)}} & {\bf 94.9\,{\scriptsize(1.0)}} & {\bf 100.0\,{\scriptsize(0.0)}} & {\bf 100.0\,{\scriptsize(0.0)}} \\
\method (log-product) & {\bf 100.0\,{\scriptsize(0.0)}} & {\bf 100.0\,{\scriptsize(0.0)}} & 90.4\,{\scriptsize(1.9)} & 68.6\,{\scriptsize(5.5)} & 84.3\,{\scriptsize(3.3)} & 99.8\,{\scriptsize(0.4)} & 90.4\,{\scriptsize(1.7)} \\
LeWM & 95.8\,{\scriptsize(1.2)} & 37.8\,{\scriptsize(2.8)} & 22.0\,{\scriptsize(9.0)} & 9.6\,{\scriptsize(6.9)} & 7.2\,{\scriptsize(3.6)} & 69.1\,{\scriptsize(12.6)} & 3.6\,{\scriptsize(2.9)} \\
\bottomrule
\end{tabular}
\end{table}
In \TRe, both methods solve nearly every instance.
As the environment becomes more complex, the success rate of LeWM drops substantially, whereas \method continues to solve most instances.
The difference persists under generalisation: without retraining, \method solves most instances in both \TRld and \TRpd, while LeWM rarely reaches the goal.
In \IS, \method with the Gödel Trick solves every instance with a maximum horizon of $T=20$.
We therefore test whether it can also find an optimal plan by setting the horizon for each instance to the length $T^*$ of its shortest plan.
\method with the Gödel Trick still solves every instance, while log-product solves $90.4\%$ and LeWM only $3.6\%$.

In \TR, instead, \method does not solve every instance.
We investigate whether these failures are associated with planning difficulty, using the length of the optimal plan as a proxy.
We group test instances by their optimal plan length and report the fraction solved in each group in \Cref{fig:planning_optimal}.
Failures concentrate on instances requiring longer plans.
LeWM mostly succeeds when the optimal plan is short and almost never when it requires more than $15$ actions.
\method remains effective at substantially longer horizons, although its success rate also decreases for the longest plans.

\paragraph{RQ4: Fuzzy semantics.}
The choice of fuzzy semantics has opposite effects on learning and planning.
During learning, only log-product grounds the concepts, whereas Gödel 
fails across all training set sizes. This can be explained by the sparse gradients discussed in \Cref{sec:method:semantics}.
Conversely, during planning, the Gödel Trick outperforms log-product (\Cref{tab:planning}), supporting the hypothesis put forward in \Cref{sec:method:semantics}.
Importantly, \method can learn with log-product and plan with the Gödel Trick, without retraining the encoder or rewriting any formula.


%% file: sections/related_works.tex
\section{Related Work}
\label{sec:related}

\paragraph{Learning world models.} \methods learn latent states from observations, relating our work to state representation learning, which learns compact states that evolve with the agent’s actions~\citep{DBLP:journals/nn/LesortRGF18}.
Such methods encode observations with a neural network and mainly differ in the training objective.
Generative world models train the encoder jointly with a decoder that reconstructs observations from the latent state~\citep{DBLP:journals/corr/abs-2301-04104, DBLP:journals/corr/abs-1803-10122}; some models also predict rewards for a given task~\citep{DBLP:conf/iclr/00010024, DBLP:journals/nature/SchrittwieserAH20}.
Like \methods, world models based on Joint Embedding Predictive Architectures (JEPA) avoid reconstruction and are trained to predict the latent encoding of the next observation~\citep{DBLP:journals/corr/abs-2306-02572}.
Their encoders are either pre-trained and frozen~\citep{DBLP:conf/icml/ZhouPLP25, DBLP:journals/corr/abs-2506-09985}, or trained jointly with the predictor using a regularization term that prevents collapse~\citep{DBLP:conf/nips/SobalZCBRL25, DBLP:journals/corr/abs-2603-19312}; \method instead constrains the latent state with logical knowledge and needs no regularisation.
Another line of work learns discrete latent states, using reconstruction losses~\citep{DBLP:journals/jair/AsaiKFM22, DBLP:conf/rlc/AgostinelliS24} or regularized prediction losses~\citep{DBLP:conf/icpr/BizzaroS26}.
Their variables, however, have no predefined meaning, which must be recovered after training by probes that map the latent to ground-truth concepts; in \method, concepts are named in advance, so the latent state needs no probe to be interpreted.
Finally, some works use LLMs to write programs describing state transitions, but these operate only on symbolic inputs and outputs~\citep{tang2024worldcoder}.


\begin{figure}[t]
  \centering
  \includegraphics[trim={.0cm .5cm .0cm .3cm}, width=0.95\linewidth]{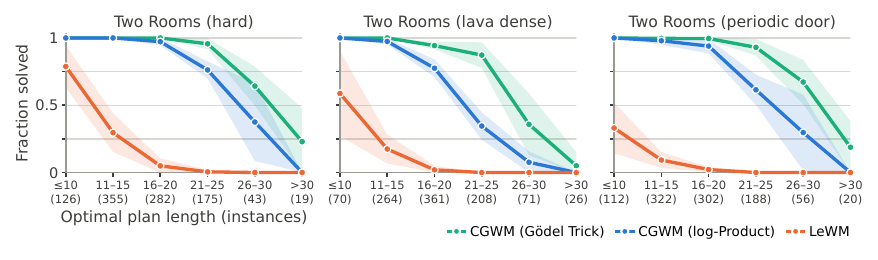}
  \caption{\textbf{\method can plan over long horizons.} Fraction of instances solved, grouped by the length of their optimal plan, with the number of instances per group in parentheses. }
  \label{fig:planning_optimal}
\end{figure}



\paragraph{Neurosymbolic world models.} Neurosymbolic learning integrates knowledge into deep learning through differentiable logics, usually under fuzzy or probabilistic semantics~\citep{DBLP:journals/ai/BadreddineGSS22, DBLP:conf/nips/ManhaeveDKDR18,pmlr-v80-xu18h}. 
Recent work extends this integration to temporal logic, where knowledge constrains sequences rather than single inputs~\citep{DBLP:conf/kr/UmiliCG23, DBLP:conf/nesy/AndreoniBDGMR25, DBLP:conf/ijcai/ManginasPR25, DBLP:journals/corr/abs-2608-16443}.

Several methods integrate knowledge into world models. 
\citet{DBLP:conf/ijcai/RamachandraSF26} inject knowledge about spatial rotations to regularise continuous latent state dynamics during training.
\citet{DBLP:journals/corr/abs-2608-17959} propose a generative world model in which additional symbolic properties, predicted from latent states, can define rewards for planning.
Closer to \method, \citet{DBLP:conf/aaai/SmetVRM25} model the environment as 
a Markov model whose symbolic state describes relations between objects, and whose transitions are probabilistic logic programs~\citep{pmlr-v216-de-smet23a}.
However, the model either receives symbolic states, or grounds initial agent positions to generate a sequence of images following an action sequence.
Neither of the latter two world models is used for planning.
\citet{DBLP:journals/corr/abs-2604-19043} learn action preconditions and effects from observations in which only the first and last symbolic states are observed. 
The predicates are given in advance together with the constraints that the action model must satisfy.
In \citet{DBLP:journals/tmlr/AhmedTBG25}, the predicates and operators of a planning domain are given as symbolic abstractions, 
and a large language model synthesizes the low-level transition functions as programs from observed transitions; observations are symbolic game states rather than images, and planning uses a classical planner over the operators.
Finally, temporal specifications are also used to define the reward of a learning agent, either by assuming symbols are given \citep{DBLP:journals/jair/IcarteKVM22, DBLP:conf/aips/GiacomoIFP19} or by learning to ground them in observations \citep{DBLP:conf/ecai/UmiliAC24}. 

%% file: sections/conclusions.tex
\section{Conclusion}
\label{sec:conclusions}

We introduced \method, a neurosymbolic approach to learning interpretable world models that support planning and robust generalisation.
As a first step, we deliberately focused on a restricted setting that isolates these aspects, assuming (i)~complete knowledge of the symbolic dynamics, (ii)~concepts that are fully observable from a single observation, and (iii)~deterministic dynamics.
These assumptions constitute limitations of the current work and motivate several directions for future research.

Hybrid latent states and predictors could combine symbolic knowledge with learned continuous components, either to represent genuinely continuous variables or to model aspects of the state and dynamics not covered by the available knowledge.
Our use of differentiable planning is deliberate in this respect: by optimising actions through backpropagation, the planner can potentially be extended to jointly optimise symbolic and continuous components.
With incomplete knowledge, an alternative is to learn the missing symbolic rules from data~\citep{DBLP:conf/ijcai/DanieleCMS23,DBLP:conf/icml/BarbieroCGZMTLP23,DBLP:conf/kr/BizzaroD26}, although weaker knowledge may introduce reasoning shortcuts~\citep{DBLP:journals/jair/MarconatoBKMUVTPT26}.
Partial observability additionally requires extending the state beyond information available in a single observation. Overall, \method demonstrates how explicit knowledge can connect interpretable representations, dynamics, and planning within a unified world model, while providing a flexible foundation for future developments.


%% file: sections/appendix_ltlf.tex
\section{Supplementary Material on \LTLF and \FLTLF}
\label{appendix:ltlf}

In this appendix, we complement the background information about \LTLF and \FLTLF presented in Sections~\ref{sec:background:LTLf} and~\ref{sec:background:fuzzy}.

\subsection{\LTLF}

As described in Section~\ref{sec:background:LTLf}, the semantics of the syntax described in~\Cref{eq:syntax} for an \LTLF formula $\varphi$ is explicitly defined over finite traces $\concepttraj=\langle \vconcepts_0,\dots,\vconcepts_{n-1}\rangle$, where the vectors $\vconcepts_t \in \{0,1\}^d$ have a Boolean value, either $0$ (\emph{false}) or $1$ (\emph{true}), for \emph{every} concept $\p\in \P$.

The syntax in~\Cref{eq:syntax} extends propositional logic on $\P$ with formulas employing the temporal operators $\tomorrow$ and $\until$, respectively representing \emph{(strong) next} and \emph{(strong) until}, which are then used as a basis for defining further operators such as $\eventually$ and $\always$.

Formally, for an \LTLF formula $\varphi$, a trace $\concepttraj$, and a time step $t \in \set{0,\ldots,n-1}$, we inductively define that $\varphi$ is true in time step $t$ of $\concepttraj$, written $\concepttraj,t\models \varphi$, as in~\citet{DBLP:conf/ijcai/GiacomoV13}: 
\begin{equation*}
\begin{array}{l l l}
\concepttraj,t \models  \p\in \P & \text{if\;\;} & c_{t, \p}=1 \\
\concepttraj,t \models  \lnot \varphi & \text{if} & \concepttraj,t \not \models \varphi\\
\concepttraj,t \models \varphi_1 \lor \varphi_2 & \text{if} & \concepttraj,t \models \varphi_1 \text{ or }  \concepttraj,t \models \varphi_2\\
\concepttraj,t \models \X \varphi &
\text{if} & t < n-1 \text{ and } \concepttraj,t+1 \models \varphi \\
\concepttraj,t \models \varphi_1 \U \varphi_2
&
\text{if } & \concepttraj,s \models \varphi_2 \text{ for some } s \text{ s.t.~} t \leq s \leq n-1 
\text{ and } \\
&  & \qquad \concepttraj,t' \models \varphi_1 \text{ for every } t' \text{ s.t.~} t \leq t' < s
\end{array}
\end{equation*}
We say that $\concepttraj$ satisfies $\varphi$, written $\concepttraj \models \varphi$, if $\concepttraj,0 \models \varphi$.

The syntax and semantics of the other standard Boolean connectives of~\Cref{eq:syntax} $\top$, $\bot$, $\land$, $\limp$ are derived as usual: $\varphi_1 \land \varphi_2 \equiv \neg(\neg \varphi_1 \lor \neg \varphi_2)$, $\varphi_1 \limp \varphi_2 \equiv \neg \varphi_1 \lor \varphi_2$, $\top\equiv\varphi \lor \lnot \varphi$, and $\bot\equiv\lnot \top$.
Further key temporal operators are derived from $\tomorrow$ and $\until$ as presented in Table~\ref{tab:ltlf-operators}.

\begin{table}[htbp]
\centering
\small
\begin{tabularx}{\linewidth}{@{}l l X@{}}
\toprule
\textbf{Operator} & \textbf{Abbreviation} & \textbf{Description} \\
\midrule
Weak next ($\Xw$) & $\Xw \varphi \equiv \X \varphi \lor \lnot \X \top$ & If there is a next instant, $\varphi$ holds at it. \\
Eventually ($\F$) & $\F \varphi \equiv \top \U \varphi$ & $\varphi$ holds at some instant of the trace. \\
Globally ($\G$) & $\G \varphi \equiv \lnot \F \lnot \varphi$ & $\varphi$ holds at every instant of the trace. \\
Release ($\release$) & $\varphi_1 \release \varphi_2 \equiv \lnot (\lnot \varphi_1 \U \lnot \varphi_2)$ & $\varphi_2$ holds up to and including the first instant at which $\varphi_1$ holds; if $\varphi_1$ never holds, $\varphi_2$ holds throughout the trace. \\
Strong release ($\mrelease$) & $\varphi_1 \mrelease \varphi_2 \equiv (\varphi_1 \release \varphi_2) \land \F \varphi_1$ & As release, but $\varphi_1$ is required to hold at some instant. \\
Weak until ($\W$) & $\varphi_1 \W \varphi_2 \equiv (\varphi_1 \U \varphi_2) \lor \G \varphi_1$ & As until, but $\varphi_2$ is not required to hold; if it never does, $\varphi_1$ holds throughout the trace. \\
\bottomrule
\end{tabularx}
\caption{Derived temporal operators of \LTLF.}
\label{tab:ltlf-operators}
\end{table}

Note that, for the sake of brevity, we often use predicate logic notation to write long \LTLF formulas in a compact manner. For example, assume we have a scenario where an agent can move on a matrix of $1,\ldots, N$ cells, but cannot stay on cells occupied by a rock. This knowledge has to be formalised as a long conjunction 
\[
(\prop{agent}(1) \limp \neg \prop{rock}(1)) \land (\prop{agent}(2) \limp \neg \prop{rock}(2)) \land \ldots \land (\prop{agent}(N) \limp \neg \prop{rock}(N))
\]
We abbreviate this long formula with the equivalent predicate logic formula 
\[
\forall g (\prop{agent}(g) \limp \neg \prop{rock}(g)) 
\]
where the \textit{for all} ($\forall$) quantifier ranges over the finite domain of $N$ cells. 
Analogously, we use the \textit{exists only one} ($\exists !$) quantifier to write formulas such as 
\[
\exists ! g (\prop{agent}(g)) 
\]
as a shorthand for $\bigvee_{i=1}^{N} \big(\prop{agent}(i) \land \bigwedge_{j\neq i} \neg \prop{agent}(j)\big)$.

\subsection{\FLTLF}

Let $\dottraj = \langle \dottensor_0, \dots, \dottensor_{n-1} \rangle$ be a fuzzy trace, where each $\dottensor_t \in [0, 1]^d$ assigns to every proposition $\p_i \in \P$ its \emph{degree of truth} $z_{t, i}$ at time step $t$.
Recall that $v(\varphi,\dottraj,t)$ denotes the degree of truth of the \FLTLF formula $\varphi$ over $\dottraj$ at time step $t$.
The fuzzy propositional connectives $\lnot$, $\land$, $\lor$, $\to$ are evaluated at each time step $t$ as follows, where $\p \in \P$ and $\varphi_1, \varphi_2$ are \FLTLF formulas:
\begin{equation*}
\begin{array}{r c l}
v(\top,\dottraj,t) & = & 1 \text{ and } v(\bot,\dottraj,t)  =  0; \\ 
v(p,\dottraj,t) & = & z_{t, p}; \\
v(\neg \varphi,\dottraj,t) & = & \tnot v(\varphi,\dottraj,t) ;\\
v(\varphi_1 \land \varphi_2,\dottraj,t) & = & v(\varphi_1,\dottraj,t)~\tand~ v(\varphi_2,\dottraj,t); \\
v(\varphi_1 \lor \varphi_2,\dottraj,t) & = & v(\varphi_1,\dottraj,t)~\tor~ v(\varphi_2,\dottraj,t) ;\\
v(\varphi_1 \rightarrow \varphi_2,\dottraj,t) & = &  v(\varphi_1,\dottraj,t)~\timpl~ v(\varphi_2,\dottraj,t) ; \\
v(\X \varphi,\dottraj,t) & = & v(\varphi,\dottraj,t+1) \text{ if } t<n-1 \text{, } 0 \text{ otherwise}; \\
v(\varphi_1 \U \varphi_2,\dottraj,t) & = &  v(\varphi_2,\dottraj,t) ~\tor~ (v(\varphi_1,\dottraj,t)~\tand~  v(\X(\varphi_1 \U \varphi_2),\dottraj,t) ).
\end{array}
\end{equation*}
where $\tand$, $\tor$, $\tnot$, $\timpl$ are the chosen t-norm, t-conorm (or s-norm), negation function and implication function, respectively. 
Note that the semantics of $\U$ (until) is given using the recursive formulation based on the expansion law for the until operator~\citep{BaiK08}.


We also consider the fuzzy counterparts of the derived \LTLF operators reported in Table~\ref{tab:ltlf-operators}, whose semantics is given explicitly below:
\begin{equation*}
\begin{array}{r c l}
v(\Xw \varphi,\dottraj,t) & = & v(\varphi,\dottraj,t{+}1) \text{ if } t< n-1 \text{, }  1 \text{ otherwise}; \\
v(\F \varphi,\dottraj,t) & = & v(\varphi,\dottraj,t) ~\tor~  v(\X~\F \varphi,\dottraj,t); \\
v(\G\varphi,\dottraj,t) & = & v(\varphi,\dottraj,t) ~\tand~ v(\Xw ~\G\varphi,\dottraj,t); \\ 
v(\varphi_1 \W \varphi_2,\dottraj,t) & = &  v(\varphi_2,\dottraj,t) ~\tor~ ( v(\varphi_1,\dottraj,t)~\tand~  v(\Xw(\varphi_1 \W \varphi_2),\dottraj,t) );  \\
v(\varphi_1 \mrelease \varphi_2,\dottraj,t) & = &  v(\varphi_2,\dottraj,t) ~\tand~ ( v(\varphi_1,\dottraj,t) ~\tor~  v(\X (\varphi_1 \mrelease \varphi_2),\dottraj,t) );  \\
v(\varphi_1 \release \varphi_2,\dottraj,t) & = &  v(\varphi_2,\dottraj,t) ~\tand~ ( v(\varphi_1,\dottraj,t) ~\tor~  v(\Xw (\varphi_1 \release \varphi_2),\dottraj,t) ). \\
\end{array}
\end{equation*}

In~\cite{DBLP:journals/corr/abs-2608-16443}, it is shown that $\F$, $\G$, $\Xw$ and $\release$ can still be obtained as abbreviations in the \FLTLF case, exactly as in Table~\ref{tab:ltlf-operators}. By contrast, $\W$ and $\mrelease$ must in general be defined explicitly. The only exception is the G\"odel semantics, where they can be derived as well.
This result also yields the characterisation of $\F$ and $\G$ given in \Cref{eq:F-G}.

%% file: sections/appendix_envs.tex
\section{Environments and Knowledge}
\label{appendix:envs}
This appendix describes the two environments and lists the knowledge given to \method, that is, the specification formulas $\Omega$, the constraints $\mathcal{K}$, and the transition rules $\mathcal{T}$ of each environment.

In this paper, we assume that the environment is deterministic and fully observable, that the size of the grid is known,
and that complete knowledge about the environment is available and exactly determines state transitions. 
We discuss extensions of this setting in \Cref{sec:conclusions}. 
The agent acts in a grid world of size $H\times W$, represented by cells $\cells=[H]\times [W]$, with $[n]:=\{1, \dots, n\}$.
At each time step $t$, the environment provides an RGB image $\bo_t \in \mathbb{R}^{lH \times lW \times 3}$, with $l$ the number of pixels along the side of a cell; and the agent executes an action $a_t\in \mathcal{A}$.
The state is described by a finite set $\concepts$ of concepts.
For each concept $\concept \in \concepts$ and cell $(h, w) \in \cells$, the proposition $\concept(\cell)$ states that concept $\concept$ holds in cell $\cell=(h, w)$.
The symbolic state at time step $t$ is the binary tensor $\vconcepts_t \in \{0,1\}^{H \times W \times |\concepts|}$ that collects the truth values of all these
propositions, and $\concepttraj = \langle \vconcepts_0, \dots, \vconcepts_T \rangle$ is the symbolic trace of the trajectory, which is not given to the model. 

All the formulas are \LTLF formulas over the propositions $\concept(\cell)$, with the quantifiers over cells introduced in \Cref{appendix:ltlf}. 
Besides the operators of \Cref{sec:background}, we use three additional ones.
The weak next $\weaknext \, \varphi$ is true at the last time step of the trace, and equals $\tomorrow \, \varphi$ at every other time step.
The proposition $\mathsf{Last}$ is true only at the last time step of the trace.
The formula $\exists! \cell.\, \varphi(\cell)$ is true when $\varphi$ is satisfied in exactly one cell.
For each action $a \in \mathcal{A}$, the proposition $a$ is true at the time steps in which the agent takes action $a$.
For an agent's movement in a direction $d$ and an integer $k\geq 1$, $\cell + k\,\delta_d$ denotes the cell $k$ steps away from $\cell$ in the direction $d$.
Propositions about cells outside the grid are false, and a constraint whose consequent refers to one specific cell $\cell + k\,\delta_d$ is quantified only over the cells for which $\cell + k\,\delta_d$ is inside the grid.
As in \Cref{sec:method:learning}, an underlined subformula $\nograd{\varphi}$ is evaluated as usual, but no gradient flows through it during training.

\subsection{\IS}
\IS is a game played on an $8\times 8$ grid of ice cells, some of which contain a rock, and one of which is the goal.
At each time step the agent chooses a direction in $\mathcal{A}_{IS}=\{\prop{N},\prop{E},\prop{S},\prop{W}\}$ and slides along it until the next cell contains a rock or is outside the grid, so it does not move if such a cell is adjacent.
Sliding over the goal does not stop the agent.
The episode ends when the agent stops on the goal, or after $20$ actions.
Rocks and goal never move.
The concepts are $\concepts_{IS} = \{\prop{agent}, \prop{rock}, \prop{goal}\}$.
The dataset contains $3200$ training and $800$ test trajectories generated by scripted policies, and about $40\%$ of the training trajectories end with the agent on the goal.
\Cref{tab:formulas_ice} lists the knowledge given to \method.
We write $\prop{at\_goal}$ for $\exists \cell.\, \prop{agent}(\cell) \wedge \prop{goal}(\cell)$, and $K = 7$ for the largest number of cells the agent can slide over with one action.
In the transition rules, $\concept'(\cell)$ is the truth value of $\concept(\cell)$ at the next time step. To simplify notation, we introduce the proposition $\prop{border}_d(\cell)$, which is true when $\cell + \delta_d$ is outside the grid.
Furthermore, to abbreviate, we write $\prop{blocked}_d(\cell)$ for $\prop{rock}(\cell + \delta_d) \vee \prop{border}_d(\cell)$, true when the next cell along direction $d$ contains a rock or is outside the grid, and $\prop{free}_{d,k}(\cell)$ for $\bigwedge_{j=0}^{k-1} \neg \prop{rock}(\cell - j\,\delta_d)$, true when $\cell$ and the $k-1$ cells before it along $d$ contain no rock. 
Formulas that contain a direction $d$ are written once for a generic direction, and the formula given to \method is the conjunction of the formulas obtained by replacing $d$ with each direction in $\mathcal{A}_{IS}$.

\begin{table}[H]
\caption{Knowledge given to \method in \IS.}
\label{tab:formulas_ice}
\centering
\small
\begin{tabular}{@{}p{0.58\linewidth}p{0.38\linewidth}@{}}
\toprule
\textbf{Formula} & \textbf{Meaning} \\
\midrule
\multicolumn{2}{@{}l}{\textit{Specification formulas} $\Omega_{IS}$} \\
$\eventually \;\prop{at\_goal}$ & The agent eventually reaches the goal. \\

\midrule
\multicolumn{2}{@{}l}{\textit{Constraints} $\mathcal{K}_{IS}$} \\
$\always(\prop{at\_goal} \rightarrow \mathsf{Last})$ & The agent is on the goal only at the last time step. \\
$\always\, \exists! \cell.\, \prop{agent}(\cell)$ & There is exactly one agent. \\
$\always\, \exists! \cell.\, \prop{goal}(\cell)$ & There is exactly one goal. \\
$\forall \cell.\, \nograd{\prop{agent}(\cell)} \rightarrow \neg \prop{rock}(\cell)$ & The agent does not start on a rock. \\
$\always\, \forall \cell.\, \big(d \wedge \weaknext\, \nograd{\prop{agent}(\cell)}\big) \rightarrow \prop{rock}(\cell + \delta_d)$ & If the agent stops in $\cell$ moving along $d$, the next cell along $d$ contains a rock. \\
$\always\, \forall \cell.\, \big(\nograd{\prop{agent}(\cell)} \wedge d \wedge \weaknext\, \neg \nograd{\prop{agent}(\cell)}\big) \rightarrow \neg \prop{rock}(\cell + \delta_d)$ & If the agent leaves $\cell$ along $d$, the next cell along $d$ does not contain a rock. \\
$\bigwedge_{k=1}^{K} (\always\, \forall \cell.\, \big(\nograd{\prop{agent}(\cell)} \wedge d \wedge \weaknext\, \nograd{\prop{agent}(\cell + k\,\delta_d)}\big) \rightarrow \bigwedge_{j=1}^{k} \neg \prop{rock}(\cell + j\,\delta_d))$ & The cells the agent slides over contain no rock. \\
$\always\, \forall \cell.\, \big(\nograd{\prop{agent}(\cell)} \wedge d \wedge \weaknext\, \nograd{\prop{at\_goal}}\big) \rightarrow \bigvee_{k=1}^{K} \prop{goal}(\cell + k\,\delta_d)$ & If the agent reaches the goal moving along $d$, the goal is on its path. \\
$\always\, \forall \cell.\, \big(\weaknext\, \nograd{\prop{at\_goal}} \wedge d \wedge \weaknext\, \nograd{\prop{goal}(\cell)}\big) \rightarrow \bigvee_{k=1}^{K} \prop{agent}(\cell - k\,\delta_d)$ & If the agent reaches the goal in $\cell$ moving along $d$, it started from a cell behind $\cell$. \\

\midrule
\multicolumn{2}{@{}l}{\textit{Transition rules} $\mathcal{T}_{IS}$} \\
$\prop{agent}'(\cell) = \bigvee_{d} \Big(d \wedge \prop{blocked}_d(\cell) \wedge \bigvee_{k=0}^{K} \big(\prop{agent}(\cell - k\,\delta_d) \wedge \prop{free}_{d,k}(\cell)\big)\Big)$ & After a movement $d$, the agent is in $\cell$ if $\cell$ is blocked along $d$ and the agent was $k$ cells before $\cell$ with no rock in between, where $k=0$ means that it did not move. \\
$\prop{rock}'(\cell) = \prop{rock}(\cell)$, \quad $\prop{goal}'(\cell) = \prop{goal}(\cell)$ & Rocks and goal do not move. \\
\bottomrule
\end{tabular}
\end{table}

\subsection{\TR}
\TR is a game played on a $9\times 15$ grid surrounded by walls and divided by an inner wall into two rooms, connected by a single door cell in the inner wall.
At each time step the agent moves by one cell in a direction in $\mathcal{A}_{TR}=\{\prop{N},\prop{E},\prop{S},\prop{W}\}$, and it does not move if the next cell is a wall or the door.
The door is closed at the beginning of the episode, opens at the time step after the agent stands on the button, and then stays open.
The episode ends when the agent reaches the goal or steps on lava, or after $40$ actions.
Walls, lava, button and goal never move.
The concepts are $\concepts_{TR} = \{\prop{agent}, \prop{goal}, \prop{wall}, \prop{lava}, \prop{button}, \prop{door}\}$.
In \TRh, the layout changes in every episode.
In the training set, about $24\%$ of the trajectories end with the agent on the goal, about $26\%$ end on the lava, and in about $62\%$ the door is open at the end.
\TRe and \TRm have no lava and a fixed inner wall in the middle of the grid.
In \TRe, the inner wall has an open passage instead of the door and there is no button, the agent starts at a random cell of the left room, and the goal is at a random cell of the right room.
\TRm adds the door, which blocks the passage until the button is pressed.
Each of these datasets contains $3200$ training and $800$ test trajectories generated by scripted policies.
\TRld is generated as \TRh, but contains $18$ lava cells instead of $6$.
\TRpd uses the layouts of \TRh and changes the dynamics. When the agent stands on the button, from the next time step the door alternates between open and closed at every time step. An agent that is on the door cell when the door closes is crushed, and the episode ends.
Both variants are test sets of $800$ instances, used only for planning.
\Cref{tab:formulas_tr} lists the knowledge given to \method in \TRh.
We write $\prop{at\_goal}$ for $\exists \cell.\, \prop{agent}(\cell) \wedge \prop{goal}(\cell)$, $\prop{on\_lava}$ for $\exists \cell.\, \prop{agent}(\cell) \wedge \prop{lava}(\cell)$, and $\prop{door\_open}$ for $\neg \exists \cell.\, \prop{door}(\cell)$, since the door disappears when it opens.
In the transition rules, $\prop{obstacle}(\cell)$ stands for $\prop{wall}(\cell) \vee \prop{door}(\cell)$, and $\prop{pressed}$ for $\exists \cell.\, \prop{agent}(\cell) \wedge \prop{button}(\cell)$.
As in \IS, the formulas written for a generic direction $d$ stand for the conjunction of the four formulas obtained by replacing $d$ with each direction in $\mathcal{A}_{TR}$.

\begin{table}[H]
\caption{Knowledge given to \method in \TRh.}
\label{tab:formulas_tr}
\centering
\small
\begin{tabular}{@{}p{0.58\linewidth}p{0.38\linewidth}@{}}
\toprule
\textbf{Formula} & \textbf{Meaning} \\
\midrule
\multicolumn{2}{@{}l}{\textit{Specification formulas} $\Omega_{TR}$} \\
$\eventually\, \prop{at\_goal}$ & The agent eventually reaches the goal. \\
$\eventually\, \prop{on\_lava}$ & The agent eventually steps on lava. \\
$\eventually\, \prop{door\_open}$ & The door is eventually open. \\

\midrule
\multicolumn{2}{@{}l}{\textit{Constraints} $\mathcal{K}_{TR}$} \\
$\always(\prop{at\_goal} \rightarrow \mathsf{Last})$ & The agent is on the goal only at the last time step. \\
$\always(\prop{on\_lava} \rightarrow \mathsf{Last})$ & The agent is on lava only at the last time step. \\
$\neg\, \prop{door\_open}$ &  At the first time step the door is closed. \\
$\always\, \exists! \cell.\, \prop{agent}(\cell)$ & There is exactly one agent. \\
$\always\, \exists! \cell.\, \prop{goal}(\cell)$ & There is exactly one goal. \\
$\always\, \exists! \cell.\, \prop{button}(\cell)$ & There is exactly one button. \\
$\always\big(\nograd{\neg\, \prop{door\_open}} \rightarrow \exists! \cell.\, \prop{door}(\cell)\big)$ & While the door is closed, it occupies exactly one cell. \\
$\forall \cell.\, \nograd{\eventually\, \prop{agent}(\cell)} \rightarrow \neg\, \prop{wall}(\cell)$ & A cell the agent occupies at some time step is not a wall. \\
$\always\, \forall \cell.\, \big(\nograd{\prop{agent}(\cell)} \wedge d \wedge \weaknext\, \nograd{\prop{at\_goal}}\big) \rightarrow \prop{goal}(\cell + \delta_d)$ & If the agent reaches the goal moving from $\cell$ along $d$, the goal is the next cell. \\
$\always\, \forall \cell.\, \big(\nograd{\prop{agent}(\cell)} \wedge d \wedge \weaknext\, \neg \nograd{\prop{at\_goal}}\big) \rightarrow \neg\, \prop{goal}(\cell + \delta_d)$ & If the agent moves from $\cell$ along $d$ without reaching the goal, the next cell is not the goal. \\
$\always\, \forall \cell.\, \big(\weaknext\, \nograd{\prop{at\_goal}} \wedge d \wedge \weaknext\, \nograd{\prop{goal}(\cell + \delta_d)}\big) \rightarrow \prop{agent}(\cell)$ & If the agent reaches the goal in $\cell + \delta_d$ moving along $d$, it was in $\cell$. \\
$\always\, \forall \cell.\, \big(\nograd{\prop{agent}(\cell)} \wedge d \wedge \weaknext\, \nograd{\prop{on\_lava}}\big) \rightarrow \prop{lava}(\cell + \delta_d)$ & If the agent steps on lava moving from $\cell$ along $d$, the next cell is lava. \\
$\always\, \forall \cell.\, \big(\nograd{\prop{agent}(\cell)} \wedge d \wedge \weaknext\, \neg \nograd{\prop{on\_lava}}\big) \rightarrow \neg\, \prop{lava}(\cell + \delta_d)$ & If the agent moves from $\cell$ along $d$ without stepping on lava, the next cell is not lava. \\
$\always\, \forall \cell.\, \big(\nograd{\prop{agent}(\cell)} \wedge d \wedge \weaknext\, \nograd{\prop{agent}(\cell)} \wedge \neg \nograd{\prop{door}(\cell + \delta_d)}\big) \rightarrow \prop{wall}(\cell + \delta_d)$ & If the agent does not move along $d$ and the next cell is not the door, the next cell is a wall. \\
$\always\, \forall \cell.\, \big(\nograd{\prop{agent}(\cell)} \wedge d \wedge \weaknext\, \neg \nograd{\prop{agent}(\cell)}\big) \rightarrow \neg\, \prop{wall}(\cell + \delta_d) \wedge \neg\, \prop{door}(\cell + \delta_d)$ & If the agent leaves $\cell$ along $d$, the next cell is neither a wall nor the closed door. \\
$\always\, \forall \cell.\, \big(\nograd{\prop{agent}(\cell)} \wedge d \wedge \neg \tomorrow\, \nograd{\prop{door\_open}} \wedge \tomorrow\tomorrow\, \nograd{\prop{door\_open}}\big) \rightarrow \prop{button}(\cell + \delta_d)$ & If the door is still closed at the next time step and open at the one after, the agent has moved onto the button. \\

\midrule
\multicolumn{2}{@{}l}{\textit{Transition rules} $\mathcal{T}_{TR}$} \\
$\prop{agent}'(\cell) = \bigvee_{d} \Big(\big(\prop{agent}(\cell - \delta_d) \wedge d \wedge \neg\, \prop{obstacle}(\cell)\big) \vee \big(\prop{agent}(\cell) \wedge d \wedge \prop{obstacle}(\cell + \delta_d)\big)\Big)$ & The agent moves by one cell along $d$, or stays in $\cell$ if the next cell is a wall or the closed door. \\
$\prop{door}'(\cell) = \prop{door}(\cell) \wedge \neg\, \prop{pressed}$ & The door opens at the time step after the agent stands on the button. \\
$\concept'(\cell) = \concept(\cell)$, for $\concept \in \{\prop{wall}, \prop{lava}, \prop{goal}, \prop{button}\}$ & Walls, lava, goal and button do not move. \\

\bottomrule
\end{tabular}
\end{table}

\Cref{tab:formulas_pd} reports the transition rules of \TRpd, which are the only part of the knowledge that changes with respect to \TRh.
Since \TRpd is only used for planning, its specification formulas and constraints are never used.
Here, we also use the operator $\mathsf{Y}$, which takes the value of $\varphi$ at the previous time step.
In the transition rules, $\prop{obstacle}'(\cell)$ stands for $\prop{wall}(\cell) \vee \prop{door}'(\cell)$, because the door is assumed to change before the agent moves.

\begin{table}[H]
\caption{Transition rules of \TRpd.}
\label{tab:formulas_pd}
\centering
\small
\begin{tabular}{@{}p{0.58\linewidth}p{0.38\linewidth}@{}}
\toprule
\textbf{Formula} & \textbf{Meaning} \\
\midrule
$\prop{door}'(\cell) = \big(\prop{door}(\cell) \wedge \prop{agent}(\cell)\big) \vee \big(\mathsf{Y}\, \prop{door}(\cell) \wedge \neg (\prop{door}(\cell) \wedge \prop{pressed})\big)$ & The door stays closed until the button is pressed, and alternates between open and closed afterwards, unless the agent is on it. \\
$\prop{agent}'(\cell) = \bigvee_{d} \Big(\big(\prop{agent}(\cell - \delta_d) \wedge \neg\, \prop{door}(\cell - \delta_d) \wedge d \wedge \neg\, \prop{obstacle}'(\cell)\big) \vee \big(\prop{agent}(\cell) \wedge \neg\, \prop{door}(\cell) \wedge d \wedge \prop{obstacle}'(\cell + \delta_d)\big)\Big) \vee \big(\prop{agent}(\cell) \wedge \prop{door}(\cell)\big)$ & The agent moves as in \TRh, with the door of the next time step blocking it, and it stays in $\cell$ forever once the door closes on it. \\
$\concept'(\cell) = \concept(\cell)$, for $\concept \in \{\prop{wall}, \prop{lava}, \prop{button}, \prop{goal}\}$ & Walls, lava, button and goal do not move. \\
\bottomrule
\end{tabular}
\end{table}

%% file: sections/appendix_results.tex
\section{Complete Experimental Results}
\label{appendix:results}

\Cref{tab:de_two_rooms,tab:de_ice} report the F1 of the concepts for every training set size, in \TRh and in \IS.
We group the concepts into static ones, which do not change position during an episode, and dynamic ones, which do.
In \TR the static concepts are $\prop{goal}$, $\prop{wall}$, $\prop{lava}$ and $\prop{button}$, and the dynamic ones are $\prop{agent}$ and $\prop{door}$.
In \IS the static concepts are $\prop{rock}$ and $\prop{goal}$, and the only dynamic one is $\prop{agent}$.
Each value is the F1 averaged over the concepts of the group, computed on the encoded states for $h{=}0$ and on the imagined trace for $h>0$.
The rows of \method are labelled with the semantics used for training.
LeWM is the version used in the main text, with one latent vector of $32$ features per cell of the grid, and LeWM (original) is its original formulation, with a single $192$-dimensional latent vector.
Since the latent vector of LeWM (original) has no grid structure, the linear probe reads the feature map of its encoder on the encoded states, and the same feature map reconstructed by a decoder from the imagined latent states.

In \IS the F1 on NeSy-MM of $\prop{agent}$ is higher on the imagined trace than on the encoded state. 
At $h=0$ the symbolic state is read from the encoder, while at $h>0$ it is obtained by sampling initial symbolic states from the encoder and applying the transition rules with the actions of the trajectory, so the knowledge of the dynamics also contributes to the estimate.
In \IS the encoder grounds the static concepts perfectly and the agent incorrectly, so the sampled states agree on the rocks and on the goal and may disagree on the position of the agent. The transition rules make the agent slide until it meets a rock or the border, and therefore map many starting positions onto the same stopping cell. Applying them to the samples concentrates the estimate, and since the true agent follows the same rules, it concentrates towards its actual position.
In \TR the same phenomenon happens with weaker intensity.
For \method and LeWM, instead, the imagined state is deterministic, and it can only inherit the errors of the encoded one.

\subsection{Ablations}
\label{appendix:results:ablations}
\Cref{tab:de_two_rooms,tab:de_ice} also report an ablation on \method trained without the entropy term $\mathcal{L}_{\text{H}}$ and without the stop-gradient on the antecedent.
Results show that both components are needed to consistently achieve a perfect grounding of the concepts. 
In \TR, no run of either ablation grounds all the concepts, and in \IS only some of the runs do. 
The standard deviation over the seeds also becomes much larger, since the runs split between those that ground the concepts and those that do not.

More precisely, we find that both techniques are primarily effective at preventing local minima. 
In the ablations, only 7 out of 10 runs on \IS ($N=3200$) with the no-entropy baseline manage to minimise the constraint losses. 
These runs have perfect grounding accuracy, like the non-ablated version. 
All other ablation runs get stuck in local minima. 

\begin{table}[!tbp]
\caption{F1 of the concepts in \TRh, for each training set size $N$ and rollout step $h$, as mean and standard deviation (in parentheses) over the seeds.}
\label{tab:de_two_rooms}
\centering
\scriptsize
\setlength{\tabcolsep}{2pt}
\begin{tabular}{rccccccccc}
\toprule
 & \multicolumn{2}{c}{Static concepts} & \multicolumn{5}{c}{Dynamic concepts} & \multicolumn{2}{c}{All concepts} \\
\cmidrule(lr){2-3}\cmidrule(lr){4-8}\cmidrule(lr){9-10}
$N$ & $h{=}0$ & $40$ & $h{=}0$ & $10$ & $20$ & $30$ & $40$ & $h{=}0$ & $40$ \\
\midrule
\multicolumn{10}{l}{\textit{\method (log-product)}} \\
3200 & 1.00 (0.00) & 1.00 (0.00) & 1.00 (0.00) & 1.00 (0.00) & 1.00 (0.00) & 1.00 (0.00) & 1.00 (0.00) & 1.00 (0.00) & 1.00 (0.00) \\
800 & 1.00 (0.00) & 1.00 (0.00) & 1.00 (0.00) & 1.00 (0.00) & 1.00 (0.00) & 1.00 (0.00) & 1.00 (0.00) & 1.00 (0.00) & 1.00 (0.00) \\
200 & 1.00 (0.00) & 1.00 (0.00) & 1.00 (0.00) & 1.00 (0.00) & 1.00 (0.00) & 1.00 (0.00) & 1.00 (0.00) & 1.00 (0.00) & 1.00 (0.00) \\
50 & 0.97 (0.07) & 0.97 (0.07) & 1.00 (0.00) & 0.98 (0.05) & 0.97 (0.08) & 0.98 (0.07) & 0.98 (0.06) & 0.98 (0.05) & 0.98 (0.07) \\
\midrule
\multicolumn{10}{l}{\textit{\method (log-product, no stop-gradient)}} \\
3200 & 0.23 (0.18) & 0.23 (0.18) & 0.30 (0.24) & 0.21 (0.17) & 0.18 (0.15) & 0.19 (0.16) & 0.19 (0.16) & 0.25 (0.11) & 0.21 (0.11) \\
800 & 0.10 (0.17) & 0.10 (0.17) & 0.40 (0.20) & 0.27 (0.14) & 0.24 (0.12) & 0.26 (0.13) & 0.26 (0.13) & 0.20 (0.10) & 0.15 (0.10) \\
200 & 0.20 (0.19) & 0.20 (0.19) & 0.25 (0.25) & 0.17 (0.17) & 0.15 (0.15) & 0.16 (0.16) & 0.16 (0.16) & 0.22 (0.13) & 0.19 (0.12) \\
50 & 0.15 (0.17) & 0.15 (0.17) & 0.15 (0.23) & 0.10 (0.16) & 0.09 (0.14) & 0.10 (0.15) & 0.10 (0.15) & 0.15 (0.14) & 0.13 (0.12) \\
\midrule
\multicolumn{10}{l}{\textit{\method (log-product, no entropy)}} \\
3200 & 0.51 (0.25) & 0.51 (0.25) & 0.85 (0.32) & 0.63 (0.33) & 0.53 (0.30) & 0.51 (0.29) & 0.49 (0.29) & 0.63 (0.27) & 0.51 (0.26) \\
800 & 0.52 (0.25) & 0.52 (0.25) & 0.93 (0.15) & 0.65 (0.23) & 0.53 (0.20) & 0.50 (0.19) & 0.48 (0.18) & 0.66 (0.21) & 0.51 (0.21) \\
200 & 0.54 (0.26) & 0.54 (0.26) & 0.90 (0.20) & 0.69 (0.27) & 0.59 (0.23) & 0.56 (0.22) & 0.54 (0.21) & 0.66 (0.24) & 0.54 (0.23) \\
50 & 0.46 (0.24) & 0.46 (0.24) & 0.90 (0.20) & 0.68 (0.19) & 0.56 (0.16) & 0.52 (0.14) & 0.50 (0.13) & 0.60 (0.22) & 0.47 (0.20) \\
\midrule
\multicolumn{10}{l}{\textit{\method (G\"odel)}} \\
3200 & 0.27 (0.16) & 0.27 (0.16) & 0.01 (0.01) & 0.01 (0.01) & 0.01 (0.01) & 0.01 (0.01) & 0.01 (0.01) & 0.18 (0.11) & 0.18 (0.11) \\
800 & 0.33 (0.15) & 0.32 (0.15) & 0.01 (0.00) & 0.00 (0.00) & 0.00 (0.00) & 0.00 (0.00) & 0.00 (0.00) & 0.22 (0.10) & 0.22 (0.10) \\
200 & 0.27 (0.17) & 0.27 (0.17) & 0.01 (0.01) & 0.00 (0.00) & 0.00 (0.00) & 0.00 (0.00) & 0.00 (0.00) & 0.18 (0.12) & 0.18 (0.12) \\
50 & 0.30 (0.19) & 0.30 (0.19) & 0.01 (0.00) & 0.00 (0.00) & 0.00 (0.00) & 0.00 (0.00) & 0.00 (0.00) & 0.20 (0.13) & 0.20 (0.13) \\
\midrule
\multicolumn{10}{l}{\textit{\method (G\"odel Trick)}} \\
3200 & 0.21 (0.10) & 0.21 (0.10) & 0.12 (0.19) & 0.08 (0.13) & 0.07 (0.12) & 0.07 (0.12) & 0.08 (0.13) & 0.18 (0.10) & 0.16 (0.08) \\
800 & 0.16 (0.10) & 0.16 (0.10) & 0.09 (0.17) & 0.06 (0.11) & 0.05 (0.10) & 0.05 (0.10) & 0.05 (0.11) & 0.13 (0.09) & 0.12 (0.08) \\
200 & 0.06 (0.08) & 0.06 (0.08) & 0.07 (0.15) & 0.05 (0.10) & 0.04 (0.09) & 0.04 (0.09) & 0.05 (0.10) & 0.06 (0.09) & 0.05 (0.07) \\
50 & 0.08 (0.08) & 0.08 (0.08) & 0.06 (0.15) & 0.04 (0.10) & 0.03 (0.09) & 0.04 (0.09) & 0.04 (0.10) & 0.07 (0.07) & 0.06 (0.06) \\
\midrule
\multicolumn{10}{l}{\textit{LeWM}} \\
3200 & 1.00 (0.00) & 1.00 (0.00) & 1.00 (0.00) & 0.90 (0.03) & 0.86 (0.04) & 0.86 (0.04) & 0.86 (0.04) & 1.00 (0.00) & 0.95 (0.02) \\
800 & 1.00 (0.00) & 1.00 (0.00) & 1.00 (0.00) & 0.80 (0.01) & 0.74 (0.02) & 0.74 (0.03) & 0.72 (0.03) & 1.00 (0.00) & 0.91 (0.01) \\
200 & 1.00 (0.00) & 1.00 (0.00) & 1.00 (0.00) & 0.72 (0.02) & 0.63 (0.01) & 0.63 (0.02) & 0.62 (0.01) & 1.00 (0.00) & 0.87 (0.00) \\
50 & 1.00 (0.01) & 1.00 (0.00) & 1.00 (0.00) & 0.70 (0.03) & 0.62 (0.03) & 0.60 (0.04) & 0.57 (0.04) & 1.00 (0.00) & 0.86 (0.01) \\
\midrule
\multicolumn{10}{l}{\textit{LeWM (original)}} \\
3200 & 0.89 (0.00) & 0.50 (0.18) & 0.90 (0.20) & 0.00 (0.00) & 0.00 (0.00) & 0.01 (0.01) & 0.01 (0.01) & 0.90 (0.07) & 0.34 (0.12) \\
800 & 0.89 (0.00) & 0.65 (0.04) & 0.65 (0.23) & 0.01 (0.01) & 0.01 (0.01) & 0.01 (0.01) & 0.01 (0.01) & 0.81 (0.08) & 0.44 (0.03) \\
200 & 0.90 (0.01) & 0.28 (0.02) & 0.85 (0.23) & 0.04 (0.01) & 0.03 (0.01) & 0.03 (0.01) & 0.03 (0.01) & 0.88 (0.08) & 0.20 (0.01) \\
50 & 0.87 (0.06) & 0.15 (0.01) & 1.00 (0.00) & 0.02 (0.00) & 0.02 (0.00) & 0.01 (0.00) & 0.01 (0.00) & 0.92 (0.04) & 0.10 (0.01) \\
\midrule
\multicolumn{10}{l}{\textit{NeSy-MM}} \\
3200 & 0.91 (0.11) & 0.91 (0.11) & 0.51 (0.01) & 0.52 (0.28) & 0.47 (0.31) & 0.43 (0.33) & 0.42 (0.33) & 0.78 (0.07) & 0.75 (0.16) \\
800 & 0.82 (0.30) & 0.82 (0.30) & 0.46 (0.15) & 0.56 (0.31) & 0.52 (0.31) & 0.50 (0.31) & 0.49 (0.31) & 0.70 (0.24) & 0.71 (0.27) \\
200 & 0.77 (0.17) & 0.76 (0.17) & 0.64 (0.21) & 0.63 (0.30) & 0.53 (0.32) & 0.48 (0.34) & 0.46 (0.35) & 0.72 (0.15) & 0.66 (0.21) \\
50 & 0.57 (0.28) & 0.57 (0.27) & 0.45 (0.15) & 0.41 (0.27) & 0.32 (0.25) & 0.26 (0.24) & 0.25 (0.24) & 0.53 (0.22) & 0.46 (0.24) \\
\bottomrule
\end{tabular}

\end{table}

\begin{table}[!tbp]
\caption{F1 of the concepts in \IS, for each training set size $N$ and rollout step $h$, as mean and standard deviation (in parentheses) over the seeds.}
\label{tab:de_ice}
\scriptsize
\centering
\setlength{\tabcolsep}{2pt}
\begin{tabular}{rccccccccc}
\toprule
 & \multicolumn{2}{c}{Static concepts} & \multicolumn{5}{c}{Dynamic concepts} & \multicolumn{2}{c}{All concepts} \\
\cmidrule(lr){2-3}\cmidrule(lr){4-8}\cmidrule(lr){9-10}
$N$ & $h{=}0$ & $20$ & $h{=}0$ & $5$ & $10$ & $15$ & $20$ & $h{=}0$ & $20$ \\
\midrule
\multicolumn{10}{l}{\textit{\method (log-product)}} \\
3200 & 1.00 (0.00) & 1.00 (0.00) & 1.00 (0.00) & 1.00 (0.00) & 1.00 (0.00) & 1.00 (0.00) & 1.00 (0.00) & 1.00 (0.00) & 1.00 (0.00) \\
800 & 1.00 (0.00) & 1.00 (0.00) & 1.00 (0.00) & 1.00 (0.00) & 1.00 (0.00) & 1.00 (0.00) & 1.00 (0.00) & 1.00 (0.00) & 1.00 (0.00) \\
200 & 1.00 (0.00) & 1.00 (0.00) & 1.00 (0.00) & 1.00 (0.00) & 1.00 (0.00) & 1.00 (0.00) & 1.00 (0.00) & 1.00 (0.00) & 1.00 (0.00) \\
50 & 1.00 (0.00) & 1.00 (0.00) & 1.00 (0.00) & 1.00 (0.00) & 1.00 (0.00) & 1.00 (0.00) & 1.00 (0.00) & 1.00 (0.00) & 1.00 (0.00) \\
\midrule
\multicolumn{10}{l}{\textit{\method (log-product, no stop-gradient)}} \\
3200 & 0.45 (0.20) & 0.45 (0.20) & 0.00 (0.00) & 0.00 (0.00) & 0.00 (0.00) & 0.00 (0.00) & 0.00 (0.00) & 0.30 (0.13) & 0.30 (0.13) \\
800 & 0.60 (0.35) & 0.60 (0.35) & 0.40 (0.49) & 0.40 (0.49) & 0.40 (0.49) & 0.40 (0.49) & 0.40 (0.49) & 0.54 (0.39) & 0.54 (0.39) \\
200 & 0.75 (0.38) & 0.75 (0.38) & 0.70 (0.46) & 0.70 (0.46) & 0.70 (0.46) & 0.70 (0.46) & 0.70 (0.46) & 0.74 (0.40) & 0.74 (0.40) \\
50 & 0.79 (0.34) & 0.79 (0.34) & 0.70 (0.46) & 0.70 (0.46) & 0.70 (0.46) & 0.70 (0.46) & 0.70 (0.46) & 0.76 (0.37) & 0.76 (0.37) \\
\midrule
\multicolumn{10}{l}{\textit{\method (log-product, no entropy)}} \\
3200 & 0.77 (0.37) & 0.77 (0.37) & 0.80 (0.40) & 0.73 (0.41) & 0.74 (0.40) & 0.73 (0.41) & 0.73 (0.42) & 0.78 (0.37) & 0.75 (0.39) \\
800 & 0.80 (0.40) & 0.80 (0.40) & 0.80 (0.40) & 0.82 (0.35) & 0.83 (0.34) & 0.83 (0.35) & 0.82 (0.36) & 0.80 (0.40) & 0.81 (0.39) \\
200 & 0.85 (0.32) & 0.85 (0.32) & 0.90 (0.30) & 0.83 (0.34) & 0.84 (0.33) & 0.83 (0.35) & 0.83 (0.35) & 0.87 (0.31) & 0.84 (0.33) \\
50 & 0.82 (0.37) & 0.82 (0.37) & 0.80 (0.40) & 0.81 (0.38) & 0.82 (0.35) & 0.82 (0.36) & 0.82 (0.36) & 0.81 (0.38) & 0.82 (0.36) \\
\midrule
\multicolumn{10}{l}{\textit{\method (G\"odel)}} \\
3200 & 0.57 (0.35) & 0.57 (0.35) & 0.00 (0.00) & 0.00 (0.00) & 0.00 (0.00) & 0.00 (0.00) & 0.00 (0.00) & 0.38 (0.23) & 0.38 (0.23) \\
800 & 0.32 (0.26) & 0.32 (0.26) & 0.00 (0.00) & 0.00 (0.00) & 0.02 (0.03) & 0.01 (0.02) & 0.00 (0.00) & 0.21 (0.18) & 0.22 (0.17) \\
200 & 0.28 (0.20) & 0.28 (0.20) & 0.00 (0.00) & 0.00 (0.00) & 0.03 (0.03) & 0.01 (0.02) & 0.00 (0.00) & 0.19 (0.13) & 0.19 (0.13) \\
50 & 0.31 (0.25) & 0.31 (0.25) & 0.00 (0.00) & 0.00 (0.00) & 0.01 (0.03) & 0.01 (0.01) & 0.00 (0.00) & 0.21 (0.16) & 0.21 (0.16) \\
\midrule
\multicolumn{10}{l}{\textit{\method (G\"odel Trick)}} \\
3200 & 0.31 (0.16) & 0.31 (0.16) & 0.47 (0.48) & 0.03 (0.04) & 0.05 (0.05) & 0.03 (0.03) & 0.06 (0.06) & 0.36 (0.25) & 0.23 (0.12) \\
800 & 0.13 (0.08) & 0.13 (0.08) & 0.00 (0.00) & 0.00 (0.00) & 0.00 (0.00) & 0.00 (0.00) & 0.00 (0.00) & 0.08 (0.05) & 0.08 (0.06) \\
200 & 0.02 (0.05) & 0.02 (0.05) & 0.00 (0.00) & 0.00 (0.00) & 0.00 (0.00) & 0.00 (0.00) & 0.00 (0.00) & 0.01 (0.04) & 0.01 (0.04) \\
50 & 0.02 (0.05) & 0.02 (0.05) & 0.00 (0.00) & 0.00 (0.00) & 0.00 (0.00) & 0.00 (0.00) & 0.00 (0.00) & 0.01 (0.04) & 0.01 (0.04) \\
\midrule
\multicolumn{10}{l}{\textit{LeWM}} \\
3200 & 1.00 (0.00) & 0.98 (0.04) & 1.00 (0.00) & 0.98 (0.02) & 0.98 (0.02) & 0.96 (0.04) & 0.95 (0.07) & 1.00 (0.00) & 0.97 (0.04) \\
800 & 1.00 (0.00) & 0.95 (0.14) & 1.00 (0.00) & 0.89 (0.12) & 0.86 (0.18) & 0.81 (0.24) & 0.78 (0.26) & 1.00 (0.00) & 0.89 (0.17) \\
200 & 1.00 (0.00) & 0.98 (0.04) & 1.00 (0.00) & 0.75 (0.02) & 0.66 (0.02) & 0.65 (0.03) & 0.65 (0.02) & 1.00 (0.00) & 0.87 (0.03) \\
50 & 1.00 (0.00) & 1.00 (0.00) & 1.00 (0.00) & 0.51 (0.05) & 0.38 (0.05) & 0.37 (0.03) & 0.37 (0.04) & 1.00 (0.00) & 0.79 (0.01) \\
\midrule
\multicolumn{10}{l}{\textit{LeWM (original)}} \\
3200 & 1.00 (0.00) & 0.63 (0.14) & 1.00 (0.00) & 0.01 (0.01) & 0.03 (0.01) & 0.02 (0.01) & 0.02 (0.01) & 1.00 (0.00) & 0.43 (0.09) \\
800 & 1.00 (0.00) & 0.69 (0.01) & 1.00 (0.00) & 0.01 (0.00) & 0.01 (0.01) & 0.01 (0.01) & 0.01 (0.01) & 1.00 (0.00) & 0.46 (0.01) \\
200 & 1.00 (0.00) & 0.55 (0.01) & 1.00 (0.00) & 0.01 (0.01) & 0.01 (0.01) & 0.02 (0.01) & 0.02 (0.01) & 1.00 (0.00) & 0.38 (0.01) \\
50 & 1.00 (0.00) & 0.42 (0.02) & 1.00 (0.00) & 0.01 (0.01) & 0.02 (0.01) & 0.02 (0.01) & 0.03 (0.02) & 1.00 (0.00) & 0.29 (0.01) \\
\midrule
\multicolumn{10}{l}{\textit{NeSy-MM}} \\
3200 & 1.00 (0.01) & 1.00 (0.01) & 0.20 (0.40) & 0.42 (0.29) & 0.52 (0.24) & 0.48 (0.26) & 0.45 (0.28) & 0.73 (0.13) & 0.82 (0.09) \\
800 & 1.00 (0.00) & 1.00 (0.00) & 0.10 (0.30) & 0.35 (0.22) & 0.46 (0.18) & 0.43 (0.19) & 0.40 (0.20) & 0.70 (0.10) & 0.80 (0.07) \\
200 & 1.00 (0.01) & 1.00 (0.01) & 0.00 (0.00) & 0.27 (0.02) & 0.40 (0.03) & 0.35 (0.03) & 0.32 (0.04) & 0.67 (0.00) & 0.77 (0.02) \\
50 & 0.99 (0.01) & 0.99 (0.01) & 0.20 (0.40) & 0.40 (0.30) & 0.49 (0.26) & 0.45 (0.28) & 0.41 (0.30) & 0.73 (0.14) & 0.80 (0.10) \\
\bottomrule
\end{tabular}

\end{table}
\newpage
\subsection{Fuzzy semantics for learning and planning}
\Cref{tab:planning_ablation} compares the modes of the two semantics, where the first and the third rows are the configurations used in the main text.
Under the log-product semantics, sampling the actions with the Gumbel-Softmax is never worse than the plain softmax, and clearly better in the harder environments.
Under the Gödel semantics, the Gödel Trick gives the same kind of improvement over the plain softmax.

\begin{table}[h]
\caption{Percentage of planning instances solved in the environment for each semantics and mode. Mean and standard deviation (in parentheses) over the seeds.}
\label{tab:planning_ablation}
\centering
\footnotesize
\setlength{\tabcolsep}{3pt}
\begin{tabular}{lccccccc}
\toprule
 & \multicolumn{5}{c}{Two Rooms} & \multicolumn{2}{c}{Ice Slider} \\
\cmidrule(lr){2-6}\cmidrule(lr){7-8}
 & easy & medium & hard & lava dense & periodic door & $T{=}20$ & $T{=}T^*$ \\
\midrule
\multicolumn{8}{l}{\textit{G\"odel}} \\
G\"odel Trick & 100.0\,{\scriptsize(0.0)} & 100.0\,{\scriptsize(0.0)} & 96.3\,{\scriptsize(0.8)} & 88.4\,{\scriptsize(2.8)} & 94.9\,{\scriptsize(1.0)} & 100.0\,{\scriptsize(0.0)} & 100.0\,{\scriptsize(0.0)} \\
softmax & 100.0\,{\scriptsize(0.0)} & 84.4\,{\scriptsize(3.4)} & 88.5\,{\scriptsize(2.6)} & 86.2\,{\scriptsize(3.5)} & 81.7\,{\scriptsize(4.1)} & 100.0\,{\scriptsize(0.0)} & 100.0\,{\scriptsize(0.0)} \\
\midrule
\multicolumn{8}{l}{\textit{log-product}} \\
Gumbel-Softmax & 100.0\,{\scriptsize(0.0)} & 100.0\,{\scriptsize(0.0)} & 90.4\,{\scriptsize(1.9)} & 68.6\,{\scriptsize(5.5)} & 84.3\,{\scriptsize(3.3)} & 99.8\,{\scriptsize(0.4)} & 90.4\,{\scriptsize(1.7)} \\
softmax & 99.9\,{\scriptsize(0.3)} & 100.0\,{\scriptsize(0.0)} & 80.6\,{\scriptsize(3.4)} & 51.7\,{\scriptsize(5.3)} & 76.9\,{\scriptsize(3.4)} & 99.4\,{\scriptsize(0.7)} & 84.8\,{\scriptsize(3.2)} \\
\bottomrule
\end{tabular}
\end{table}



%% file: sections/appendix_implementation.tex
\section{Implementation Details}
\label{appendix:implementation}
This appendix describes the architectures, the losses and the hyperparameters of \method and of the baselines.
The environment is a grid of $H\times W$ cells $\cells$, $\concepts$ is the set of concepts, and $z_{t,\cell,\concept}$ is the degree of truth that concept $\concept$ holds in cell $\cell$ at time step $t$. A trajectory has at most $T_{\max}$ actions, $40$ in \TR and $20$ in \IS.

\subsection{Implementation Details of \method}
\label{appendix:implementation:cgwm}
\paragraph{Encoder.}
The observations are RGB images, with three colour channels and $l=8$ pixels per cell, so an image of \TR has $72\times 120$ pixels and an image of \IS has $64\times 64$, with values rescaled to $[0,1]$.
The encoder applies three convolutional layers with $2\times 2$ kernels, stride $2$, and $32$, $64$ and $64$ channels, each followed by a ReLU.
The three strides reduce each $8\times 8$ block of pixels to a single position, so the output is one $64$-dimensional feature vector per cell, computed only from the pixels of that cell.
Two $1\times 1$ convolutions, the first with $32$ channels and a ReLU and the second with $|\concepts|$ channels, map each feature vector to one logit per concept, and the sigmoid function $\sigma$ turns each logit into a degree of truth in $[0,1]$.
\paragraph{Losses.}
We consider a batch of $B$ trajectories, where trajectory $b$ has $T_b$ actions $a^b_0, \dots, a^b_{T_b-1}$, labels $y^b_1, \dots, y^b_M$, and fuzzy trace $\dottraj^b = \langle \dottensor^b_0, \dots, \dottensor^b_{T_b} \rangle$.
Trajectories shorter than $T_{\max}$ are padded, and the padded time steps are ignored by every term.
Under the log-product semantics, the four terms of \Cref{eq:loss} are
\begin{align*}
    \mathcal{L}_{\text{spec}} &= -\frac{1}{BM} \sum_{b=1}^{B} \sum_{m=1}^{M} \Big[ y^b_m \log v(\omega_m, \dottraj^b) + (1-y^b_m) \log\big(1-v(\omega_m, \dottraj^b)\big) \Big], \\
  \mathcal{L}_{\text{cons}} &= -\frac{1}{B\,|\mathcal{K}|} \sum_{b=1}^{B} \sum_{\kappa \in \mathcal{K}} \log v(\kappa, \dottraj^b), \\
   \mathcal{L}_{\text{pred}} &= \frac{1}{\sum_{b} T_b} \sum_{b=1}^{B} \sum_{t=0}^{T_b-1} \mathrm{BCE}\big(\hat{\dottensor}^b_{t+1},\, \dottensor^b_{t+1}\big),
  \qquad \hat{\dottensor}^b_{t+1} = \Pred(\dottensor^b_t, a^b_t), \\
  \mathcal{L}_{\text{H}} &= -\frac{1}{|\concepts|\,|\cells| \sum_{b} (T_b+1)} \sum_{b=1}^{B} \sum_{t=0}^{T_b} \sum_{\concept \in \concepts} \sum_{\cell \in \cells} \mathrm{H}\big(z^b_{t,\cell,\concept}\big).
\end{align*}
The binary cross-entropy is averaged over all the transitions of the batch, the cells and the concepts, and $\mathrm{H}(x) = -x\log x - (1-x)\log(1-x)$ is the binary entropy.
The weight of the entropy term at epoch $e$ is $\lambda_{\text{H}}(e) = \lambda^0_{\text{H}} \max\big(0,\, 1 - e/\lceil E/2 \rceil\big)$, where $E$ is the number of epochs and $\lambda^0_{\text{H}} = 1.5$, so it decreases linearly to zero during the first half of training.

\paragraph{Training.}
We use Adam \citep{DBLP:journals/corr/KingmaB14} with learning rate $3\cdot 10^{-4}$, and batches of $32$ trajectories.
The number of epochs is chosen so that every run performs the same number of optimisation steps whatever the training set size, $6{,}400$ in \TR and $3{,}200$ in \IS, up to rounding.

\paragraph{Planning.}
For each instance, \method optimises $128$ plans in parallel.
The entries of the matrix $U$ of each plan (\Cref{sec:method:planning}) are drawn from $\mathcal{N}(0, 0.1^2)$ and optimised with Adam, with learning rate $0.1$, for $300$ steps.
Under the log-product semantics, the degrees of truth of the actions are sampled with the Gumbel-Softmax \citep{DBLP:conf/iclr/JangGP17}, $A_t = \mathrm{softmax}(\mathbf{u}_t + \mathbf{g}_t)$, where the entries of $\mathbf{g}_t$ are drawn from a standard Gumbel distribution at every step, and the loss is $-\log v\big(\psi, \hat{\dottraj}(A)\big)$.
Under the Gödel semantics, the Gödel Trick perturbs the plan in the same way, $\tilde{\mathbf{u}}_t = \mathbf{u}_t + \mathbf{g}_t$, and sets $A_{t,a} = \sigma\big(\tilde{u}_{t,a} - \tfrac{1}{2}(\tilde{u}^{(1)}_t + \tilde{u}^{(2)}_t)\big)$, where $\tilde{u}^{(1)}_t$ and $\tilde{u}^{(2)}_t$ are the two largest entries of $\tilde{\mathbf{u}}_t$.
Only the action with the largest perturbed value has a degree of truth above $1/2$, and the loss is $1 - v\big(\psi, \hat{\dottraj}(A)\big)$.

\subsection{Implementation Details of Baselines}
\label{appendix:implementation:baselines}
\paragraph{LeWM.}
The encoder of LeWM has the same three convolutional layers as the encoder of \method, followed by a $1\times 1$ convolution with $32$ output channels and batch normalisation, so that the latent state is a $32$-dimensional vector per cell.
The predictor receives the latent state and the action, repeated in every cell, and applies a $3\times 3$ convolution with $64$ channels followed by residual $3\times 3$ convolutional blocks, two in \TR and six in \IS, and a $1\times 1$ convolution whose output is added to the input latent state.
Following \citet{DBLP:journals/corr/abs-2603-19312}, the prediction loss is the mean squared error between the predicted and the encoded latent states, and SIGReg~\citep{DBLP:journals/corr/abs-2511-08544} is applied to the latent vectors of the cells, with $64$ random projections and weight $0.1$.
The reward head is a multilayer perceptron with $256$ hidden units that maps the latent state to a scalar reward, and is trained with the squared error.
The reward is $1$ when the agent is on the goal, and, in \TR, $-1$ when the agent is on lava and $0.5$ at the first time step in which the agent stands on the button.
LeWM is trained with the same optimiser, learning rate, batch size and gradient clipping as \method, for $25{,}600$ optimisation steps in \TR and $12{,}800$ in \IS.

To plan, LeWM encodes the current image and optimises $128$ plans in parallel, initialised and optimised as those of \method, for $300$ steps.
The objective of a plan is the sum of the rewards predicted along its imagined trace.
LeWM then executes the first five actions of the plan with the highest objective and observes the new image.
It repeats the same procedure every five actions until $T_{\max}$ actions have been executed: it encodes the last image, optimises the remaining part of the plan for $50$ steps, starting from the previous solution, and executes its next five actions.
The original LeWM plans with the cross-entropy method, which is designed for the continuous action spaces of its control tasks, while here an action is one of four directions. 
We therefore optimise its plans with a gradient-based procedure, as for \method, so that the two planners differ only in their objective.
We also let LeWM re-observe the environment every five actions, while \method plans once from the first image.
This is motivated by \Cref{tab:de_two_rooms,tab:de_ice}, since the concepts that a linear probe recovers from the imagined states of LeWM degrade as the rollout grows, so re-planning from a fresh observation keeps every imagined rollout short. This removes part of the error accumulated when imagining.

Since the latent state of LeWM does not explicitly contain the concepts, we read them with a linear probe, a $1\times 1$ convolution that maps the latent vector of each cell to one value per concept.
The probe is trained with the binary cross-entropy against the true concepts of the encoded states of the first $300$ training trajectories of the run, for $\min(300,N)$ epochs (where $N$ is the number of trajectories) with Adam and a learning rate of $0.01$. The probe is then applied to both the encoded and the imagined latent states.

\paragraph{LeWM (original).}
LeWM (original) follows the original formulation, in which the latent state is a single $192$-dimensional vector.
The encoder has the same three convolutional layers, followed by a fully connected layer with $256$ units, batch normalisation, a ReLU, and a linear layer with $192$ outputs.
The predictor adds to the latent vector the output of a multilayer perceptron with $512$ hidden units, which receives the latent vector and the one-hot action.
It is trained with the same losses as LeWM, with SIGReg applied to the latent vectors with $1024$ random projections, and with the same reward head, optimiser and number of steps.
Since its latent vector has no grid structure, a decoder, consisting of a multilayer perceptron with $256$ hidden units, is trained to reconstruct the feature map of the encoder from the latent vector, with no gradient flowing back into the world model.
The probe then reads the $64$-dimensional feature vector of each cell. On the encoded states it reads the feature map of the encoder, and on the imagined states it reads the feature map reconstructed by the decoder.
With this latent state, the probe recovers the dynamic concepts from the encoded states, but its F1 on the imagined states falls close to zero already after a few imagined steps (\Cref{tab:de_two_rooms,tab:de_ice}).
A plan optimised on such rollouts carries no information about where the agent is, which is why the version used in the main text keeps one latent vector per cell.

\paragraph{NeSy-MM.}
NeSy-MM receives the same concepts, transition rules, constraints and specification formulas as \method, and the same trajectory labels. \method encodes an image into a fuzzy symbolic state and evaluates every formula once on the resulting fuzzy trace, obtaining a degree of truth, while NeSy-MM uses its perception to define a probability distribution over crisp symbolic states, and evaluates the same formulas with Boolean semantics on states sampled from it.
The perception of NeSy-MM has the same architecture as the encoder of \method, and outputs a categorical distribution over the cells for $\prop{agent}$ and $\prop{goal}$, and for $\prop{button}$ in \TR, and a Bernoulli distribution per cell for the other concepts.
Given an observation, NeSy-MM samples $100$ symbolic states from the distribution provided by the encoder. The transition rules are then applied to each sampled state, with the crisp actions of the trajectory, and since the environment is deterministic and the rules are complete, this results in a crisp symbolic trace.
Each constraint and each specification formula is then evaluated on this trace under classical semantics, and is therefore either true or false.
NeSy-MM is trained by maximising the log-probability that the constraints are evaluated as true and that the specification formulas match their respective labels. This is estimated over the $100$ samples.
The gradient with respect to the sampled concepts is estimated with the RLOO estimator~\citep{kool2019buy}.
NeSy-MM is trained with Adam, batches of $32$ trajectories and the same number of optimisation steps as \method, with learning rate $10^{-3}$ in \TR and $3\cdot 10^{-4}$ in \IS.
To measure the grounding, for each concept we take the most probable cell if the concept occupies exactly one cell, and all the cells with probability above $1/2$ otherwise.

\subsection{Evaluation Protocol}
\label{appendix:implementation:protocol}
Every model is trained with the ten seeds and with $N \in \{50, 200, 800, 3200\}$ training trajectories.
For a given seed, the $N$ trajectories are the first $N$ of a random permutation of the $3200$ training trajectories, so the smaller training sets are contained in the larger ones.
The seed also determines the initialisation of the model, the order of the batches, and the noise sampled during training for the model that requires it.
During planning, for each seed, $100$ instances are sampled without replacement from the $800$ test trajectories of the environment, and all the methods plan on the same instances.
\method plans with the world model trained with the log-product semantics and the same seed on $3200$ trajectories of \TRh or of \IS.
LeWM plans with the model trained with the same seed on the corresponding environment, which also includes \TRe and \TRm.
An instance counts as solved when executing the plan in the environment brings the agent to the goal.
In the setting $T{=}T^*$ of \IS, $T^*$ is the length of the shortest plan of the instance, computed by breadth-first search.

\subsection{Hardware and Reproducibility}
\label{appendix:implementation:reproducibility}
\paragraph{Hardware.}
All experiments were run on a machine with two Intel Xeon Silver 4410Y processors, $256$\,GB of RAM, and two NVIDIA RTX A6000 GPUs with $48$\,GB of memory each.
\paragraph{Reproducibility.}
We do not distribute the datasets, as they are generated procedurally and deterministically by the code we release.
Each dataset is generated from a fixed seed. All the other sources of randomness are seeded as well.
The code, together with the configuration file of every run and the versions of the libraries, is available at \url{https://anonymous.4open.science/r/cgwm-726E}.